\documentclass[11pt]{article}

\usepackage{microtype}
\usepackage{graphicx}
\usepackage{subfigure}
\usepackage{booktabs} 

\usepackage{algorithm,algorithmic}

\usepackage{natbib}
\usepackage{fullpage}
\usepackage[utf8]{inputenc}
\usepackage[T1]{fontenc}
\usepackage{microtype}
\usepackage{csquotes}
\usepackage[osf,sc]{mathpazo}
\RequirePackage[scaled=0.90]{helvet}
\RequirePackage[scaled=0.85]{beramono}
\RequirePackage{textcomp}
\usepackage{xcolor}
\definecolor{DarkGreen}{rgb}{0.075,0.375,0.075}
\definecolor{DarkRed}{rgb}{0.5,0.1,0.1}
\definecolor{DarkBlue}{rgb}{0.1,0.1,0.5}
\definecolor{Gray}{rgb}{0.2,0.2,0.2}
\usepackage[pdftex]{hyperref}
\hypersetup{
    unicode=false,          
    pdftoolbar=true,        
    pdfmenubar=true,        
    pdffitwindow=false,      
    pdfnewwindow=true,      
    colorlinks=true,       
    linkcolor=DarkBlue,          
    citecolor=DarkGreen,        
    filecolor=DarkRed,      
    urlcolor=DarkBlue,          
    pdftitle={},
    pdfauthor={},
}
\usepackage[skip=2mm,indent=5mm]{parskip}
\usepackage[small]{caption}
\RequirePackage{amsthm,amsmath,amsfonts,amssymb}

\usepackage[capitalize,noabbrev]{cleveref}

\usepackage[utf8]{inputenc} 
\usepackage[T1]{fontenc}    
\usepackage{url}            
\usepackage{booktabs}       
\usepackage{amsfonts}       
\usepackage{nicefrac}       
\usepackage{microtype}      
\usepackage{xcolor}         
\usepackage{amsmath}
\usepackage{amsthm} 
\usepackage{enumitem}
\usepackage{pifont}
\usepackage{wrapfig}
\usepackage{bm}
\usepackage{bbm}
\usepackage{todonotes}
\usepackage{algorithm}
\usepackage{algorithmic}

\def\eqref#1{equation~\ref{#1}}

\def\1{\bm{1}}

\def\vg{{\bm{g}}}

\def\vs{{\bm{s}}}

\DeclareMathAlphabet{\mathsfit}{\encodingdefault}{\sfdefault}{m}{sl}
\SetMathAlphabet{\mathsfit}{bold}{\encodingdefault}{\sfdefault}{bx}{n}

\def\gC{{\mathcal{C}}}
\def\gD{{\mathcal{D}}}
\def\gE{{\mathcal{E}}}
\def\gF{{\mathcal{F}}}

\def\gO{{\mathcal{O}}}

\def\gS{{\mathcal{S}}}

\DeclareMathOperator*{\argmin}{arg\,min}

\title{How Benchmark Prediction from Fewer Data \\ Misses the Mark}

\author{%
  Guanhua Zhang\footnote{Corresponding author: \href{mailto:guanhua.zhang@tuebingen.mpg.de}{guanhua.zhang@tuebingen.mpg.de}}{~}$^{1,2}$, Florian E. Dorner$^{1,2,3}$, Moritz Hardt$^{1,2}$
}
\date{
$^{1}$\textit{Max Planck Institute for Intelligent Systems, Tübingen}\\
$^{2}$\textit{Tübingen AI Center}\\
$^{3}$\textit{ETH Zurich}\\
}

\newcounter{daggerfootnote}
\newcommand*{\daggerfootnote}[1]{%
    \setcounter{daggerfootnote}{\value{footnote}}%
    \renewcommand*{\thefootnote}{\fnsymbol{footnote}}%
    \footnote[2]{#1}%
    \setcounter{footnote}{\value{daggerfootnote}}%
    \renewcommand*{\thefootnote}{\arabic{footnote}}%
    }

\begin{document}

\onecolumn
\maketitle

\begin{abstract}
Large language model (LLM) evaluation is increasingly costly, prompting interest in methods that speed up evaluation by shrinking benchmark datasets. Benchmark prediction (also called efficient LLM evaluation) aims to select a small subset of evaluation points and predict overall benchmark performance from that subset. In this paper, we systematically assess the strengths and limitations of 11 benchmark prediction methods across 19 diverse benchmarks. First, we identify a highly competitive baseline: Take a random sample and fit a regression model on the sample to predict missing entries. Outperforming most existing methods, this baseline challenges the assumption that careful subset selection is necessary for benchmark prediction. Second, we discover that all existing methods crucially depend on model similarity. They work best when interpolating scores among similar models. The effectiveness of benchmark prediction sharply declines when new models have higher accuracy than previously seen models. In this setting of extrapolation, none of the previous methods consistently beat a simple average over random samples. To improve over the sample average, we introduce a new method inspired by augmented inverse propensity weighting. This method consistently outperforms the random sample average even for extrapolation. However, its performance still relies on model similarity and the gains are modest in general. This shows that benchmark prediction fails just when it is most needed: at the evaluation frontier, where the goal is to evaluate new models of unknown capabilities\daggerfootnote{Code is available at \url{https://github.com/socialfoundations/benchmark-prediction}.}.
\end{abstract} 
\section{Introduction}
\label{sec:intro}

Increasingly, computational cost is a major bottleneck in the evaluation of recent generative models. Growing model size and benchmark task difficulty, as well as the sheer number of available benchmarks all escalate the problem. For example, evaluating a single 176B parameter model on the HELM multi-task benchmark required 4,200 GPU hours \cite{Liang2023HolisticEO}; even major companies noted the significant computational burden of evaluation on the BigBench multi-task benchmark \cite{ganguli2023challenges}.

The problem has prompted much recent work on more efficient LLM evaluation. The typical approach is to find a subset of data points to evaluate on, and to predict benchmark performance from these few evaluations. The simplest method is the \emph{random sample mean}: Take a random sample of $n$ evaluation points, and compute the mean of the benchmark metric on the sample. For a metric, like accuracy, with values in the interval $[0,1]$, the sample mean gives an additive approximation up to error $O(1/\sqrt{n}).$ More sophisticated methods try to improve over this baseline by following a common strategy: cleverly choose a small \emph{core set} of evaluation points, evaluate multiple known models on these points, then fit a model to predict overall benchmark performance from these evaluations.

We group existing efforts following this strategy under the term \emph{benchmark prediction}. 
Previous research has proposed several hypotheses for why benchmark prediction can work: Core sets identify the most informative data points~\citep{Rodriguez2021EvaluationEA}, they exploit the dependence between model performance on different data points~\citep{anchorpoints}, and they can capture the unobserved abilities of models~\citep{tinybench}. 

The goal of our work is to systematically examine the strengths and weaknesses of benchmark prediction as a solution concept for efficient LLM evaluation.

\begin{figure}[t]
    \centering
    \includegraphics[width=0.95\linewidth]{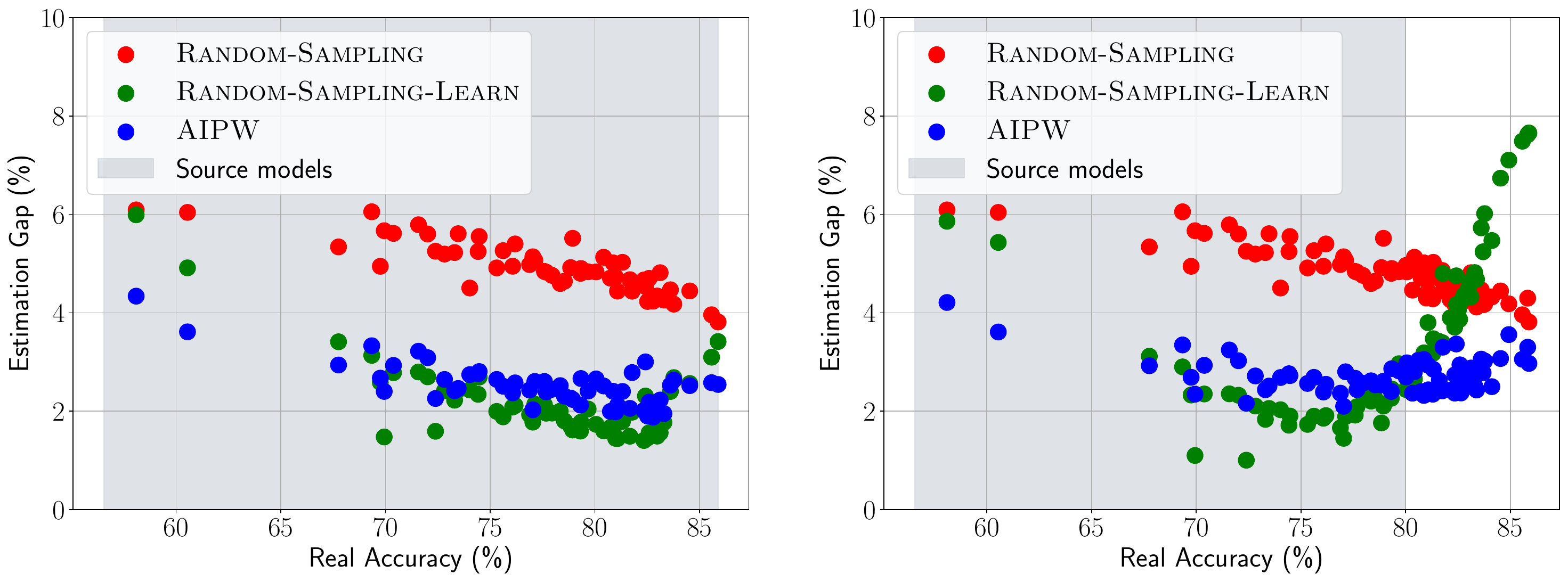}
    \caption{
    The $x$-axis denotes the real accuracy in \texttt{ImageNet} while the $y$-axis denotes the estimation gap (\eqref{eq:problem}) for each target model. 
    The gray stands for the accuracy range of source models.
    Left: source models are randomly sampled across all models.
    Right: source models are randomly sampled from models with accuracy lower than 80\%.
    }
    \label{fig:banner}
\end{figure}

\subsection{Our Contributions}
We conduct a large-scale, systematic evaluation of 11 state-of-the-art benchmark prediction methods across 19 diverse benchmarks. 
For each benchmark, we collect detailed performance results for at least 83 models on all data points. 
We split all models into two groups: source models and target models. 
For the source models, performance data is available for all data points. 
In contrast, for the target models, performance information is available only for up to 50 data points.
Each method must adhere to this constraint of selecting no more than 50 data points, and the objective is to estimate each target model’s mean performance across the full benchmark. 
To evaluate method effectiveness, we compute the {average estimation gap}—the absolute difference between the true and estimated full-benchmark performances across all target models.

\paragraph{Many methods work well on similar models, but a simple baseline works best.}
We first study the \emph{interpolation} regime where the source and target models are random drawn from the same set of all models. Our empirical findings first confirm that in this regime it is possible to reduce the {average estimation gap} relative to \textsc{Random-Sampling}, which simply reports the mean performance on a randomly selected core set. 
All evaluated methods outperform \textsc{Random-Sampling} in over half of the benchmarks. Given that all methods operate on the same number of core-set evaluations, their computational costs are comparable.  Thus, a lower \textit{average estimation gap} indicates superior data efficiency—effectively, more informative use of the evaluation budget.

Surprisingly, what works best in the interpolation regime is remarkably simple: after randomly sampling a core set, rather than computing its mean, we fit a regression model to predict the true mean performance. 
This method, \textsc{Random-Sampling-Learn}, consistently outperforms most other methods and reduces the \textit{average estimation gap} by an average of 37\% compared to \textsc{Random-Sampling}. 
This suggests that the manner of core-set selection is relatively unimportant; rather, the key to success is modeling the correlation between core-set and full-benchmark performances.

\paragraph{Methods fail at the evaluation frontier.}
However, our analysis reveals a major limitation: the effectiveness of benchmark prediction methods drops sharply when source and target models are \emph{not} drawn from the same distribution. We call this the \emph{extrapolation} regime. To explore this regime, we conduct an experiment in which we select the top-performing models (according to the full benchmark) as target models and use the poorer-performing ones as source models. This setup reflects the typical use of benchmarking at the \emph{evaluation frontier} where new models are being released that are likely better than existing ones. 

In the extrapolation regime, we show that most benchmark predictions methods fail to outperform the naive \textsc{Random-Sampling} baseline. This is illustrated in Figure~\ref{fig:banner}. When source models cover the full range of performances (left), \textsc{Random-Sampling-Learn} more than halves the estimation error. 
When source models are restricted to the lower-than-80\% accuracy (right), \textsc{Random-Sampling-Learn} still outperforms \textsc{Random-Sampling} for target models similar to the source distribution, but its predictions substantially degrade for better-performing targets outside the source range.

\paragraph{AIPW is an overlooked exception to the rule.}
One notable exception is a method inspired by augmented inverse propensity weighting (\textsc{AIPW})---used in other statistical applications---that we introduce in the context of benchmark prediction. 
\textsc{AIPW} reliably outperforms \textsc{Random-Sampling} both under interpolation and extrapolation. 
Although it sometimes performs worse than 
\textsc{Random-Sampling-Learn} when targets resemble sources, it consistently maintains an advantage when they do not, thanks to being a consistent estimator. 
However, as illustrated in Figure~\ref{fig:banner} (right), even \textsc{AIPW} sees diminishing improvements as target models’ accuracies exceed those of the sources.

\paragraph{Benchmark prediction relies on model similarity.}
To more systematically examine the generalization of benchmark prediction methods, we calculate the \emph{model similarity}~\citep{Mania2019ModelSM}, quantifying how closely the predictions of each target model match those of the source models used in training.
We observe a strong negative correlation between model similarity and estimation gap: methods that beat \textsc{Random-Sampling} tend to do so primarily for targets similar to sources, while accuracy on disimilar models deteriorates. 
In contrast, \textsc{Random-Sampling} exhibits neutral correlation, providing consistent (albeit less accurate) estimates regardless of similarity.

\paragraph{Main takeaway.}
Our findings suggest that while benchmark prediction techniques can be useful in specific scenarios, their reliance on similarity between source and target models poses a risk of misestimating the performance of new models. 
This underscores the importance of applying these methods with caution, especially for evaluating models that significantly deviate from previous ones.

\section{Related Work}
Evaluating large language models (LLMs) has become increasingly costly as these models grow in size and capabilities~\citep{Liang2023HolisticEO,open-llm-leaderboard-v2,Yu2023SkillMixAF,Zhou2024OnSU}. 
These costs manifest in several ways. 
First, the collection and annotation of evaluation data can require significant resources~\citep{Yue2023AutomaticEO}. 
To mitigate these costs, researchers have turned to methods such as using LLMs-as-judges~\citep{Gu2024ASO,Hackl2023IsGA} or employing active labeling~\citep{Kossen2022ActiveSE,Kossen2021ActiveTS,corneanu2020computing,deng2021labels,zouhar2025selectdatapointsefficienthuman} to generate evaluation data and labels. 
However, these savings come with drawbacks. 
For instance, LLM-as-a-judge does not produce reliable evaluation outcomes, as judge models tend to prefer models similar to them, and have other biases ~\citep{Wataoka2024SelfPreferenceBI,Panickssery2024LLMER,Dorner2024LimitsTS,chen2024humans}.

Another significant cost in LLM benchmarking arises from the model inference itself.
Generating responses with LLMs can be time-consuming~\citep{Liang2023HolisticEO,Zhou2024OnSU,shi2024best}, and common inference time scaling techniques~\citep{sun2020test,hardt2023test,snell2024scaling,lewis2020retrieval} may exacerbate this issue. 
The success of scaling laws~\citep{kaplan2020scaling,Ruan2024ObservationalSL} in predicting model performance has fueled interest in the development of benchmark prediction techniques~\citep{anchorpoints,tinybench,polo2024efficient,Owen2024HowPI,Pacchiardi2025PredictaBoardBL}, which aim to estimate benchmark performance by evaluating LLMs on a limited set of data
\footnote{Unlike bandit literature~\citep{Zhou2024OnSU,shi2024best}, which focuses on identifying the best model from a pool, benchmark prediction is more challenging as it seeks to forecast overall benchmark performance for any new model.}.

The key idea underpinning benchmark prediction is that not all evaluation examples carry the same amount of information~\citep{Rodriguez2021EvaluationEA}. 
It is hypothesized that a smaller core set of examples can represent the entire test set, allowing for accurate estimation of overall benchmark performance~\citep{anchorpoints}. This is similar to efficient model training approaches, which aim to identify a subset of training data that enable performance comparable to training on the full dataset~\citep{Sachdeva2023DataDA,Zhou2023ProbabilisticBC}. 
Indeed, a popular benchmark prediction method, k-medoids clustering, is a classical approach to core-set selection for training~\citep{Farahani2009FacilityLC}. However, it is important to recognize that the objectives of training and evaluation differ significantly. 
While training focuses on minimizing empirical risk and enhancing model performance, evaluation seeks to provide an unbiased estimation of a model’s performance to facilitate fair model comparison~\citep{Owen2024HowPI}.
Our work challenges the assumption that core-set selection is the key to the success of benchmark prediction by introducing competitive methods that do not rely on core-set selection.

Many existing approaches treat benchmark prediction as a learning problem, aiming to predict a model’s overall performance based on its performance on a subset of data~\citep{anchorpoints,tinybench,Li2024ActiveEA,metabench,sortandsearch}. 
Despite promising results, previous work has highlighted limitations in terms of estimation variance ~\citep{Madaan2024QuantifyingVI}. Going further, we highlight that most benchmark prediction methods rely on model similarity, with estimation performance deteriorating when target models deviate from familiar source models.

\section{What is Benchmark Prediction?}

\subsection{Problem Formulation}
We define a benchmark as a triplet $(\gD, \gF, s)$. Here $\gF$ refers to the set of models to be evaluated on the benchmark and $s$ represents the evaluation metric. Lastly, $\gD$ represents the benchmark data with $|\gD|=N$ data points. A data point is referred to as $z \in \gD$, where $z=(x,y)$, $x$ refers to the query and $y$ refers to the ground truth answer. 

    \begin{itemize}[left=2pt]
        \item $s(f, z)$ refers to the performance of any $f \in \gF$ on any data point $z \in \gD$.  For example, $s(f, z)=\mathbbm{1}[f(x)=y]$ if the benchmark uses standard accuracy as the metric. 
        \item $\bar{s}(f, \gD')=\frac{1}{|\gD'|} \sum_{z \in \gD'} s(f, z)$ represents the average performance of $f \in \gF$ on any $\gD' \subset \gD$.
        \item $\vs(f, \gD')=\{s(f, z)\}_{z \in \gD'}$ represents the vectorized performance of $f \in \gF$ on all data points in $\gD' \subset \gD$, and $\vs(\gF', z)=\{s(f, z)\}_{f \in \gF'}$ represents the vectorized performances of all models in $\gF' \subset \gF$ on data point $z \in \gD$.
        \item $S(\gF', \gD')=\{\vs(f, \gD')\}_{f \in \gF'}=\{\vs(\gF', z)\}^\texttt{T}_{z \in \gD'}$ is the performance matrix of all models in $\gF' \subset \gF$ on all data points in $\gD' \subset \gD$.
    \end{itemize}
    We refer to $\gF^{(s)} =\{f_{1}, \ldots, f_{M}\} \subset \gF$ as the set of source models, whose performances on every data point of the benchmark $S(\gF^{(s)}, \gD)$ are known. The rest of the models are the target models $\gF^{(t)}=\gF \setminus \gF^{(s)}$, which are only be evaluated on $n \ll N$ data points to save computational costs.

Benchmark prediction with fewer data aims to estimate $\bar{s}(f, \gD)$ for every $f \in \gF^{(t)}$ with only $n$ data points.
In practice, benchmark prediction often involves two steps:  \ding{172} identifying a representative core-set $\gC \subset \gD$ with $|\gC|=n$ data points, and \ding{173} learning a performance estimator $h$ to estimate the average performance on the full benchmark based on the core-set.
Formally, the goal of benchmark prediction  is to find $\gC$ and $h$ to minimize the estimation gap over target models,
\begin{align}
    \text{estimation gap:~~~~~~}
   \frac{1}{|\gF^{(t)}|} \sum_{f \in \gF^{(t)}} \big| \bar{s}(f, \gD) - h\big[ \vs(f, \gC), S(\gF^{(s)}, \gD) \big]  \big|\,\text{.}
   \label{eq:problem}
\end{align}
For simplicity, in the remainder of the paper, we will denote the estimated performance of target model $f\in\gF^{(t)}$ as $h(f)$, instead of explicitly writing $h[ \vs(f, \gC), S(\gF^{(s)}, \gD)]$.

\subsection{Benchmark Prediction Methods}
\paragraph{Previous methods.}
In this paper, we examine five widely-used benchmark prediction methods,
\begin{itemize}[left=4pt]
    \item \textsc{Random-Sampling} randomly samples a subset as $\gC$ and directly returns the mean performances as $h^{\textsc{Random-Sampling}}(f)$.
    \item \textsc{Anchor-Points-Weighted}~\citep{anchorpoints} uses k-medoids clustering to identify $\gC$ and returns a weighted sum based on the density of each cluster as $h^{\textsc{Anchor-Points-Weighted}}(f)$.  
    \item \textsc{Anchor-Points-Predictor}~\citep{anchorpoints} extends \textsc{Anchor-Points-Weighted}.
    Instead of directly returning the weighted sum, a linear regression model is learned as $h^{\textsc{Anchor-Points-Predictor}}(f)$.
    \item \textsc{P-IRT}~\citep{tinybench} extends \textsc{Anchor-Points-Predictor} by replacing the regression model with an Item Response Theory (IRT) model as $h^{\textsc{P-IRT}}(f)$. 
    \item \textsc{GP-IRT}~\citep{tinybench} further generalizes $\textsc{P-IRT}$ by combining its estimation with \textsc{Anchor-Points-Weighted} as a weighted sum, and use it as $h^{\textsc{GP-IRT}}(f)$.
\end{itemize}
\paragraph{New methods.}
We introduce six methods that have not yet been applied to benchmark prediction.
\begin{itemize}[left=4pt]
    \item \textsc{Random-Sampling-Learn} randomly samples a subset as $\gC$ and learns a Ridge regression model $g$, which predict $\bar{s}(f, \gD)$ based on $\vs(f, \gC)$, as $h^{\textsc{Random-Sampling-Learn}}(f)$.
    \item \textsc{Random-Search-Learn} performs \textsc{Random-Sampling-Learn} for 10,000 times and selects the run based on cross-validation. 
    \item \textsc{Lasso} trains a Lasso regression model to predict $\bar{s}(f, \gD)$ based on $\vs(f, \gD)$ with sparsity constraints on number of non-zero weights lower than $n$.
    The learned model is then used as $h^{\textsc{Lasso}}(f)$.
    \item \textsc{Double-Optimize} employs gradient descent to optimize both a subset selection vector, which models $\gC$, and a linear regression model, $h^{\textsc{Double-Optimize}}$~\citep{Jang2016CategoricalRW,Bengio2013EstimatingOP}.
    \item Principal Component Analysis (\textsc{PCA}) treats benchmark prediction as a matrix completion problem by assuming the performance matrix $S(\gF, \gD)$ is of low rank. 
    By randomly sampling a subset as $\gC$, this methods conducts PCA to impute the missing values for target models~\citep{Vershynin2016FourLO,cai2010singular}.
    \item Augmented inverse propensity weighting (AIPW) \cite{robins1995semiparametric}: Inspired by the application of prediction powered inference ~\citep{Angelopoulos2023PPIEP,Angelopoulos2023PredictionpoweredI} to the LLM-as-a-judge setting \cite{boyeau2024autoeval,Dorner2024LimitsTS}, we apply a more general AIPW estimator to benchmark prediction. 
    We train a Ridge regression model $g$ for every target model $f$, which predicts the point-wise performance $s(f,z)$ based on $\vs(\gF^{(s)}, z)$.
    Formally, 
    \begin{align}
        g = \argmin_{g'} \frac{1}{n} \sum_{z \in \gC} \big[
        g'[\vs(\gF^{(s)}, z)] - s(f, z)
        \big]^2 \text{.}
    \end{align}
    The idea behind the AIPW estimator is to use the predicted performance $\hat{s}(f, z)=g[\vs(\gF^{(s)}, z)]$ as a proxy score to estimate $\bar{s}(f,\mathcal{D})$ and ``debias'' that estimator as follows
    \begin{align}
    h^{\textsc{AIPW}}(f) = \bar{s}(f, \gC) + \frac{1}{1+\frac{n}{N-n}}\left(\frac{1}{N-n} \sum_{z \in \gD-\gC} \hat{s}(f, z)\ - \frac{1}{n} \sum_{z \in \gC} \hat{s}(f, z)\right) \text{.}
    \end{align}
    Unlike the other learning-based baselines, AIPW is a consistent estimator for $\bar{s}(f, \gD)$\cite{glynn2010introduction}. Compared to \textsc{Random-Sampling}, it reduces estimator variance by a factor of up to $\frac{1}{1+\frac{n}{N}} \rho(\hat{s}(f,z),s(f,z))^2$ \cite{Dorner2024LimitsTS}, where $\rho$ is the Pearson correlation coefficient.
\end{itemize}
More details of each method are listed in Appendix~\ref{app:method}. 
\section{Experiments}
\label{sec:exp}

\subsection{Experiment Setup}
We select a diverse range of benchmarks from the following sources\footnote{Since \textsc{P-IRT} and \textsc{GP-IRT} requires $s(f, z)$ to be binary, we only use benchmarks with accuracy as metric.}.
\begin{itemize}[left=0pt]
    \item {HELM-Lite} benchmarks~\citep{Liang2023HolisticEO}: \texttt{OpenbookQA}~\citep{openbookqa}, \texttt{GSM8K}~\citep{gsm8k}, \texttt{LegalBench}~\citep{legalbench}, \texttt{Math}~\citep{math}, \texttt{MedQA}~\citep{MedQA}, and \texttt{MMLU}~\citep{mmlu}. 
    We obtain the per-data point performances of $|\gF|=83$ models from the official leaderboard.  
    \item {GLUE} benchmarks~\citep{Wang2018GLUEAM}: \texttt{MRPC}~\citep{mrpc}, \texttt{RTE}~\citep{rte,rte2,rte3}, \texttt{SST-2}~\citep{sst2}, \texttt{MNLI}~\citep{mnli}, and \texttt{QNLI}~\citep{qnli}. We use the per-data performances of $|\gF|=87$ models provided by AnchorPoint\footnote{The provided score file for \texttt{QQP} is broken so we exclude it.}~\citep{anchorpoints}.
    \item {OpenLLM} benchmarks~\citep{open-llm-leaderboard-v2}: \texttt{IFEval}~\citep{ifeval}, \texttt{Math}~\citep{math}, \texttt{MMLU-Pro}~\citep{mmlupro}, \texttt{Arc-Challenge}~\citep{allenai:arc}, \texttt{BBH}~\citep{bbh}, \texttt{GPQA}~\citep{gpqa} and \texttt{MUSR}~\citep{musr}. We use $|\gF|=448$ models provided by Huggingface~\footnote{\url{https://huggingface.co/spaces/open-llm-leaderboard/open_llm_leaderboard\#}} and collect their performance scores. 
    \item \texttt{ImageNet}~\citep{imagenet}: We collect $|\gF|=110$ models from Pytorch Hub~\footnote{\url{https://pytorch.org/vision/stable/models.html\#classification}} and evaluate them on \texttt{ImageNet}.
\end{itemize}
A summary of benchmark statistics is provided in Appendix~\ref{app:setup}.

\subsection{Estimation Gap Reduction under Interpolation}
As done in previous work~\citep{anchorpoints,tinybench}, we examine the effectiveness of benchmark prediction methods under the interpolation model split where source models are identically distributed with target models.

\paragraph{Interpolation model split.} For each benchmark, we randomly select \( 75\% \) of models as source models $\gF^{(s)}$, for which performance scores across all data points $S(\gF^{(s)}, \gD)$ are available. 
The remaining \( 25\% \) of models serve as target models $\gF^{(t)}$ for assessment of benchmark prediction methods. 
Each target model is evaluated on only $n=50$ data points unless specified otherwise.
Benchmark prediction methods are used to estimate the full benchmark average performance $\bar{s}(f, \gD)$ of each target model $f \in \gF^{(t)}$ and evaluated based on the estimation gap from \eqref{eq:problem}.
Each experiment is repeated over 100 random trials, and we report the average estimation gap across all target models in these trials to ensure robustness. See standard errors in Appendix~\ref{app:results}.

\begin{figure}[t]
    \centering
    \includegraphics[width=0.99\linewidth]{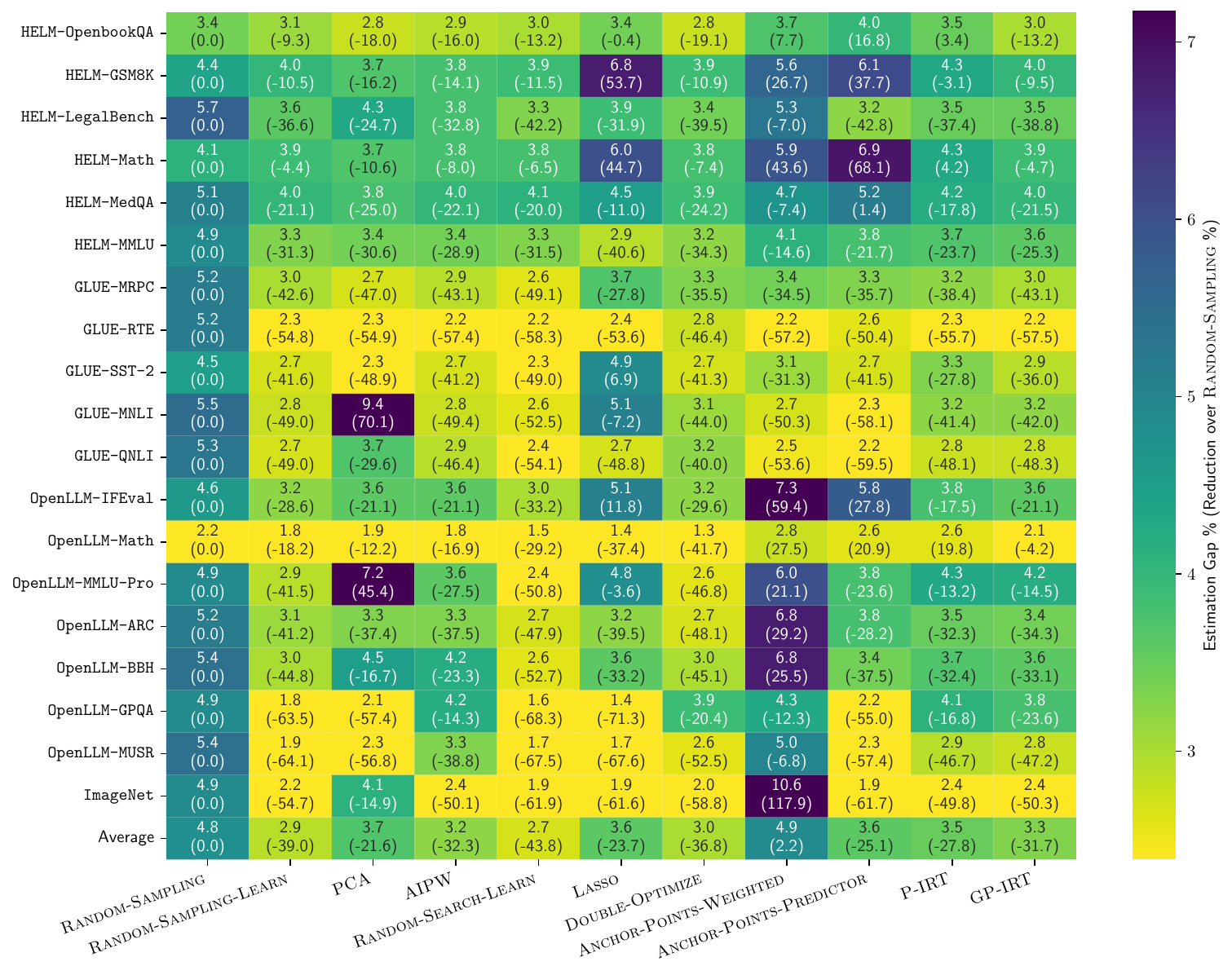}
    \caption{
    The estimation gaps ($\downarrow$) for target models (\eqref{eq:problem}) under the interpolation split, where source and target models are identically distributed.
    Each target is evaluated on $n=50$ data points.
    The estimation gap reduction ($\downarrow$) over \textsc{Random-Sampling} is shown in parentheses. 
    A negative reduction means that the method achieves a lower gap than \textsc{Random-Sampling}.
        \% is omitted.
     }
    \label{fig:acc_gap}
\end{figure}

\paragraph{Results.} The results are presented in Figure~\ref{fig:acc_gap}. 
Compared to \textsc{Random-Sampling}, all other benchmark methods effectively reduce the estimation gap in over half of the evaluated benchmarks. 
Notably, nine out of ten methods reduce the estimation gap by more than 20\% on average across all benchmarks, as indicated in the last row. 
This verifies the effectiveness of benchmark prediction methods in the interpolation setting, where source and target models are identically distributed.
Interestingly, the top-performing method is the simple baseline, \textsc{Random-Search-Learn}, which achieves a 42.1\% reduction compared to \textsc{Random-Sampling} averaged accross all benchmarks. 
In comparison to the previous state-of-the-art, $\textsc{GP-IRT}$, which leads to a 29.9\% reduction on average, \textsc{Random-Search-Learn} results in a lower estimation gap in nearly all benchmarks.

On the other hand, the selection of the core-set does not significantly enhance the effectiveness of benchmark prediction. 
For example, the second best-performing method, \textsc{Random-Sampling-Learn}, also consistently outperforms \textsc{Random-Sampling} across all benchmarks, despite the sole difference being the use of a Ridge regression model rather than directly averaging across the core-set. 
With a 37.2\% reduction in estimation gap, it performs comparably to \textsc{Random-Search-Learn}, despite the latter conducting 10,000 iterations of \textsc{Random-Sampling-Learn} to identify the best subset. 
Moreover, it surpasses methods like \textsc{Double-Optimize} and \textsc{GP-IRT}, which select subsets through optimization or clustering. 
Another benchmark prediction method, \textsc{AIPW}, which also utilizes a randomly sampled core-set, consistently achieves a lower estimation gap across all benchmarks, yielding results comparable to the state-of-the-art \textsc{GP-IRT}. 
These findings challenge the prevailing notion that the core of benchmark prediction lies in identifying the most informative or representative subset. 
Instead, our results suggest that the primary driver of benchmark prediction success is learning to predict the mean, with core-set selection playing a relatively minor role.

\begin{figure}[t]
    \centering
    \includegraphics[width=0.99\linewidth]{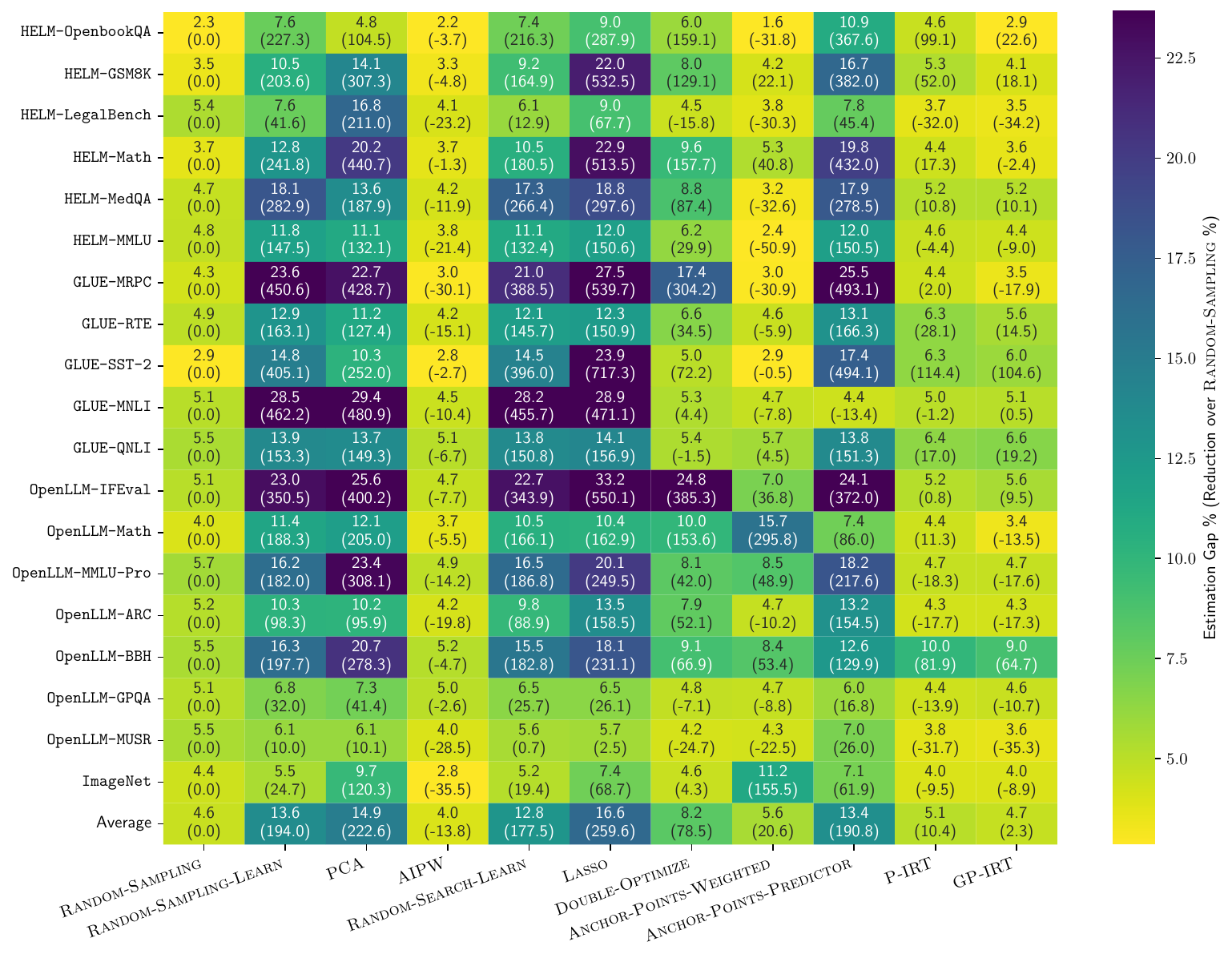}
    \caption{
    The estimation gaps ($\downarrow$) for target models (\eqref{eq:problem}) under extrapolation split, where source models are the lowest-performing 50\%, and target models are the top 30\%.
    Each target model is evaluated on $n=50$ data points.
    We also report the estimation gap reduction ($\downarrow$) over \textsc{Random-Sampling} in parentheses. 
    A negative reduction implies that the method achieves a lower estimation gap than \textsc{Random-Sampling}.
        \% is omitted.
     }
    \label{fig:acc_gap_top}
\end{figure}

\subsection{Estimation Gap Increase under Extrapolation}
We examine the effectiveness of benchmark prediction methods under the extrapolation model split where target models all perform better than source models.
\paragraph{Extrapolation model split.} Different from the random source-target model split last subsection, we begin by ranking all models for a given benchmark based on their average performance on the full benchmark $\bar{s}(f, \gD)$. 
The lowest-performing 50\% of these models are designated as source models, while the top 30\% serve as target models for evaluating benchmark prediction methods.  
This strategy reflects real-world model development scenarios, where developers debug and assess improved models based on existing less effective models.
The estimation gap as defined in \eqref{eq:problem} is used for measuring the effectiveness of benchmark prediction. We again repeat each experiment 100 times.

\paragraph{Results.}
The results are shown in Figure~\ref{fig:acc_gap_top}.
The average estimation gap for \textsc{Random-Sampling} (4.6\%) is largely comparable with the interpolation setting (4.8\%) as it doesn't rely on source models.
However, for all other methods, the estimation gap increases when compared to the interpolation setting. 
Nearly all benchmark prediction methods that outperform \textsc{Random-Sampling} in the interpolation scenario now show diminished performance.
Notably, the previous best method \textsc{Random-Search-Learn} now results in a 185.1\% increase in estimation gap than \textsc{Random-Sampling}, and performs worse than \textsc{Random-Sampling} across all benchmarks.
The only method that still outperforms \textsc{Random-Sampling} on average is \textsc{AIPW}, beating \textsc{Random-Sampling} in all 19 benchmarks.
This is because \textsc{AIPW}, like \textsc{Random-Sampling}, is a consistent estimator, but has lower variance than \textsc{Random-Sampling} when its predictor is effective.
However, the estimation gap reduction (-13.8\%) of \textsc{AIPW} over \textsc{Random-Sampling} in the extrapolation setting is also less pronounced than in the interpolation setting (-32.3\%).

This stark contrast between interpolation and extrapolation settings underscores the heavy reliance of most benchmark prediction methods on the similarity between source and target models. 
This is unsurprising, given that many methods approach benchmark prediction as a machine learning problem, which often struggles in out-of-domain scenarios. 
However, unlike traditional machine learning, which primarily emphasizes in-domain performance, a key objective of benchmarking is to assess and identify new superior models. 
Therefore, extrapolation is a more prevalent and pertinent setting than interpolation in the context of benchmarking, and the decline in the estimation gap of benchmark prediction methods in this setting calls for more caution.

\subsection{Reliance on Model Similarity}
In this subsection, we investigate the extent to which benchmark prediction methods rely on the similarity between target and source models.
\paragraph{Model similarity.}
We follow previous works~\citep{Mania2019ModelSM,Goel2025GreatMT} and define the model similarity of target model $f$ to all source models $\gF^{(s)}$ as follows, 
\begin{align}
    & \gS(f, \gF^{(s)}, \gD) = \frac{1}{M} \sum\nolimits_{f' \in \gF^{(s)}} \frac{c_{obs} - c_{exp}}{1 - c_{exp}}\,\text{.} \label{eq:sim} 
\end{align}
Here, $c_{exp}=\bar{s}(f, \gD)\bar{s}(f', \gD) + (1-\bar{s}(f, \gD))(1-\bar{s}(f', \gD))$ measures the chance agreement rate, \emph{i.e.}, the expected probability of $\{s(f,z)=s(f',z)\}$ if $s(f,z)$ is independent of $s(f',z)$. 
In contrast, $c_{obs}=\frac{1}{N} \sum_{z \in \gD} \mathbbm{1}[s(f, z)=s(f',z)]$ is the observed agreement rate.
For simplicity, we use $\gS(f)$ to denote $\gS(f, \gF^{(s)}, \gD)$ in the remainder of the paper.
$\gS(f)$ quantifies how similar the performance pattern of the target model $f$ is to all source models $\gF^{(s)}$, with a higher value indicating greater similarity~\citep{Goel2025GreatMT}.

\begin{figure}[t]
    \centering
    \includegraphics[width=0.9\linewidth]{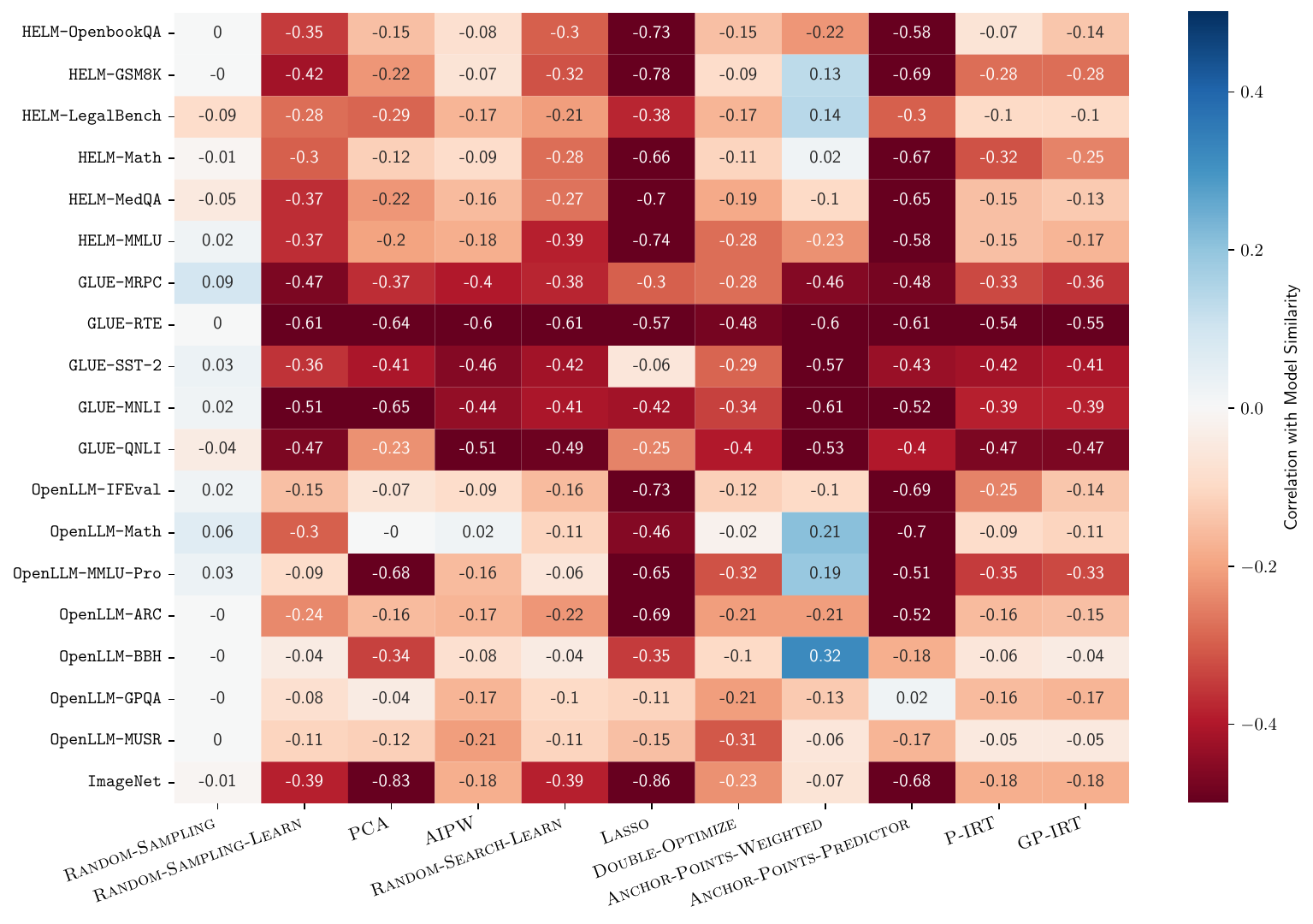}
    \caption{The Pearson correlation between normalized per-model estimation gap (\eqref{eq:norm_gap}) and model similarity (\eqref{eq:sim}). Negative correlation indicates that target models that are dissimilar to source models tend to have larger estimation gap, and vice versa.}
    \label{fig:corr_sim}
\end{figure}

We aim to examine the correlation between model similarity and estimation gap.
However, we note that the estimation depends on the standard deviation of $s(f,z)$.
Since we use accuracy as the metric in our experiment,  $s(f, z)$ is Bernoulli with parameter $p_f=\bar{s}(f, \gD)$ and standard deviation $\sigma_f=\sqrt{p_f(1-p_f)}$.
By randomly sampling $n$ data points as $\gC$, Chebyshev's inequality ensures that
\begin{align}
    |\bar{s}(f, \gC) - \bar{s}(f, \gD)| < {\sigma_f}/{\sqrt{\alpha n}}
\end{align}
with probability at least $(1-\alpha)$.
In other words, the performance of target models with lower $\sigma_f$ is easier to estimate with the same amount of data. Thus, the standard deviation of the basic estimation gap could potentially confound the observed correlation between model similarity and estimation gap. 
Consider the method \textsc{Random-Sampling}, whose estimation does not depend on source models. 
If all target models with low $\sigma_f$ coincidentally have high $\gS(f)$, while those with high $\sigma_f$ have low $\gS(f)$, then a spurious correlation between estimation gap and model similarity to target models could appear even for \textsc{Random-Sampling}. To prevent this, we define the normalized estimation gap as 
\begin{align}
\text{normalized estimation gap for $f$:~~~~~} \gE(f) &= \frac{1}{\sigma_f}  | \bar{s}(f, \gD) - h(f)  |\,\text{.} \label{eq:norm_gap}
\end{align}
Then we measure the Pearson correlation between model similarity in \eqref{eq:sim} and the normalized estimation gap in \eqref{eq:norm_gap}.

\paragraph{Results.}

The results are shown in Figure~\ref{fig:corr_sim}.
A clear negative correlation between model similarity and estimation gap emerges for almost all benchmark prediction methods except for \textsc{Random-Sampling}. In particular, the best-performing method under the interpolation model split, \textsc{Random-Sampling-Learn}, exhibits a negative correlation below -0.2 in 13/19 benchmarks. Despite its asymptotic unbiasedness, we also find negative correlations for \textsc{AIPW}. This is perhaps unsurprising: While \textsc{AIPW} is consistent independent of how well its regression model $g[\vs(\gF^{(s)},z)]$ predicts $s(f, z)$, its variance depends precisely on that prediction quality. If the predictions are good, \textsc{AIPW} improves substantially over \textsc{Random-Sampling}, while there is no improvement when predictions are fully uninformative. 
But intuitively, predicting $s(f, z)$ is harder when $f$ is very different from the models $\gF^{(s)}$ used for training the predictor $g[\vs(\gF^{(s)},z)]$.

\subsection{Ablation on Core-set Size}
We conduct an ablation study on the size of the core-set $n$.
We experiment with $n=$ $10$, $20$, $50$, $100$, and $200$, and the summarized results are shown in Table~\ref{tab:ablation} (detailed results can be found in Appendix~\ref{app:results}). 
As expected, the estimation gap generally decreases as $n$ increases for most methods. 
Our previous conclusions remain valid across both settings. 
With larger core-set sizes, most methods continue to perform better than \textsc{Random-Sampling} in the interpolation split but fail to do so in the extrapolation model split. 
Interestingly, we also find that \textsc{Random-Sampling} outperforms all other methods when given twice as much data, even in the interpolation split.

\begin{figure}[t]
    \centering
    \includegraphics[width=0.42\linewidth]{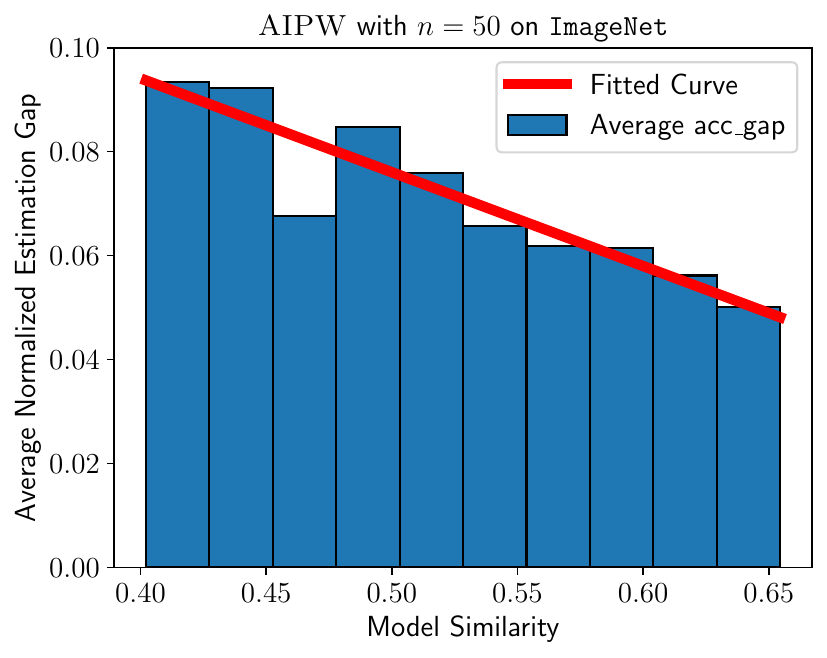}
    \includegraphics[width=0.42\linewidth]{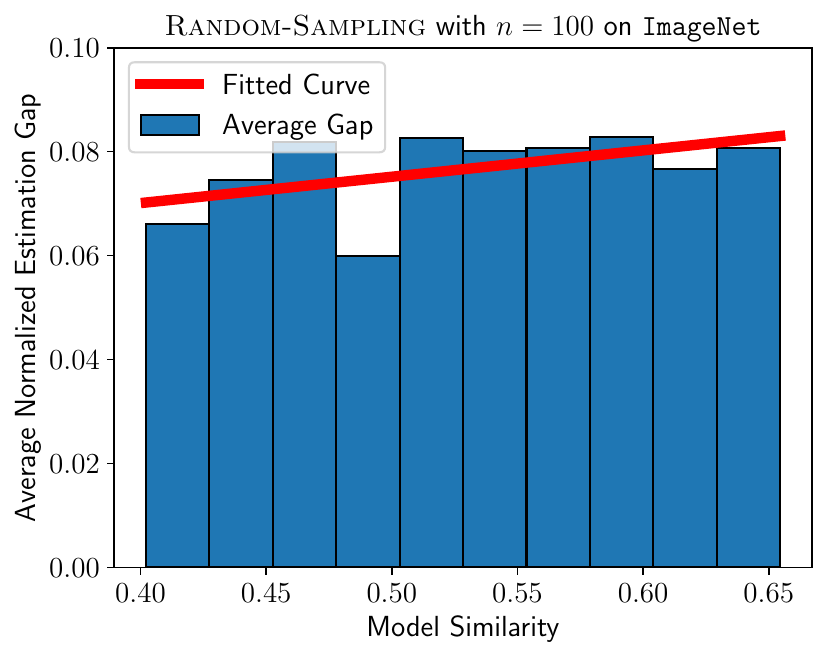}
    \caption{
    Average normalized estimation gap relative to model similarity for \textsc{AIPW} (n=50) and \textsc{Random-Sampling} (n=100) on \texttt{ImageNet}. Each bar represents the target models whose similarity to source models falls within the corresponding range. The normalized estimation gap is defined as shown in \eqref{eq:norm_gap}. On average, \textsc{AIPW} outperforms \textsc{Random-Sampling}, even with half the data. However, \textsc{Random-Sampling} shows better performance when model similarity is low.
    }
    \label{app_fig:aipw50_compare_rand100}
\end{figure}

\textsc{AIPW} remains effective in both settings.
Its advantage over \textsc{Random-Sampling} remains stable as $n$ increases.
While \textsc{AIPW} reduces the estimation gap by -32.3\% in interpolation and -13.8\% in extrapolation for $n=50$, the reductions remain -31.3\% and -12.8\%, respectively, for $n=200$.
This is consistent with its variance reduction depending on prediction quality rather than vanishing as $n$ increases.
Figure~\ref{app_fig:aipw50_compare_rand100} compares \textsc{AIPW} with $n=50$ to \textsc{Random-Sampling} with $n=100$ data points using \texttt{ImageNet}. 
\textsc{AIPW} achieves a lower average normalized estimation gap compared to \textsc{Random-Sampling}, despite using only half the data.
However, the normalized estimation gap for \textsc{AIPW} is biased with respect to model similarity. 
In contrast, the normalized estimation gap under \textsc{Random-Sampling} remains largely neutral regarding model similarity. 
Consequently, while \textsc{AIPW} reduces the average, it produces a higher gap for models with low similarity compared to \textsc{Random-Sampling} with twice the data.

\begin{table}[t]
\caption{
Ablation study on the core-set size $n$.
We report the estimation gap averaged over all benchmarks.
\% is neglected for each metric.
The lowest estimation gap in each column is highlighted in bold.
See detailed results in Appendix~\ref{app:results}.
}
\label{tab:ablation}
\centering
\resizebox{\textwidth}{!}{
\begin{tabular}{c|ccccc|ccccc}
\toprule
\midrule
&  \multicolumn{5}{c|}{Interpolation} & \multicolumn{5}{c}{Extrapolation} \\
& \multicolumn{1}{c|}{$n=10$}
& \multicolumn{1}{c|}{$n=20$}
& \multicolumn{1}{c|}{$n=50$}
& \multicolumn{1}{c|}{$n=100$}
& \multicolumn{1}{c|}{$n=200$}
& \multicolumn{1}{c|}{$n=10$}
& \multicolumn{1}{c|}{$n=20$}
& \multicolumn{1}{c|}{$n=50$}
& \multicolumn{1}{c|}{$n=100$}
& \multicolumn{1}{c}{$n=200$}\\
\midrule
\midrule
\textsc{Random-Sampling} & 11.0 & 7.7 & 4.8 & 3.3 & 2.1 & 10.7 & 7.4 & 4.6 & 3.1 & 2.0 \\
\textsc{Random-Sampling-Learn} & 5.4 & 4.2 & 2.9 & 2.1 & 1.5 & 17.6 & 15.8 & 13.6 & 12.1 & 11.1 \\
\textsc{PCA} & 6.6 & 5.2 & 3.7 & 2.8 & 2.1 & 19.9 & 17.6 & 14.9 & 12.2 & 9.3 \\
\textsc{AIPW} & 8.3 & 5.4 & 3.2 & 2.2 & 1.5 & \textbf{9.6} & \textbf{6.5} & \textbf{4.0} & \textbf{2.7} & \textbf{1.8} \\
\textsc{Random-Search-Learn} & \textbf{4.5} & \textbf{3.7} & \textbf{2.7} & \textbf{2.0} & {1.4} & 16.0 & 14.4 & 12.8 & 11.8 & 11.1 \\
\textsc{Lasso} & 7.8 & 6.1 & 3.6 & 2.6 & 2.2 & 22.0 & 19.3 & 16.6 & 15.3 & 14.6 \\
\textsc{Double-Optimize} & 6.6 & 4.8 & 3.0 & 2.3 & 1.9 & 11.3 & 9.0 & 8.2 & 8.0 & 7.0 \\
\textsc{Anchor-Points-Weighted} & 8.9 & 6.9 & 4.9 & 4.0 & 3.2 & 10.4 & 6.7 & 5.6 & 4.7 & 3.4 \\
\textsc{Anchor-Points-Predictor} & 4.7 & 4.1 & 3.6 & 3.4 & 4.1 & 16.2 & 14.8 & 13.4 & 12.4 & 11.2 \\
\textsc{P-IRT} & 7.3 & 5.9 & 3.5 & 2.1 & \textbf{1.3} & 9.8 & 8.2 & 5.1 & 3.7 & 3.0 \\
\textsc{GP-IRT} & 7.2 & 5.7 & 3.3 & 2.1 & 1.4 & 9.7 & 7.8 & 4.7 & 3.4 & 2.5 \\
\midrule
\bottomrule
\end{tabular}}
\end{table}

\section{Conclusion}
In this paper, we study the problem of benchmark prediction from fewer data and examine 11 benchmark prediction methods.
Our findings call into question the necessity of meticulous core-set selection and reveal that these methods are most proficient at interpolating scores among similar models. 
However, except \textsc{Random-Sampling} and \textsc{AIPW}, all methods face significant difficulties when predicting target models that differ substantially from those they have encountered before.

We caution against the indiscriminate use of benchmark prediction techniques, as their dependence on model similarity causes most of them to fail precisely when most needed: at the evaluation frontier, where the aim is to assess new models with unknown capabilities. 
Even in the context of interpolation, no method outperforms \textsc{Random-Sampling}, when that simple baseline is given access to twice as much data.
Thus, while we recommend to use \textsc{AIPW} as a consistent estimator with lower variance, this suggests that simply raising the sampling budget for \textsc{Random-Sampling} can be competitive, especially in settings where predictions of other models for fitting \textsc{AIPW} are costly to obtain.

\paragraph{Acknowledgement.}
We thank Yatong Chen and Jiduan Wu for helpful discussions and feedback on draft versions of this work.
Florian Dorner is grateful for financial support from the Max Planck ETH Center for Learning Systems (CLS).

\newpage
\bibliographystyle{plain}
\bibliography{ref}
\newpage
\appendix
\onecolumn
\section{Details of Benchmark Prediction Methods}
\label{app:method}

\subsection{Problem Formulation}
We repeat the notation and the problem formulation here for the reader's convenience.
\begin{itemize}[left=2pt]
    \item  A benchmark is represented as a triplet $(\gD, \gF, s)$.
    \item $\gD$ represents the benchmark data with $|\gD|=N$ data points. A data point is referred to as $z \in \gD$, where $z=(x,y)$, $x$ refers to the query and $y$ refers to the ground truth answer.
    \item $\gF$ refers to all potential models that can be evaluated on the benchmark.
    \item $s$ represents the metric of the benchmark. 
    \begin{itemize}
        \item $s(f, z)$ refers to the performance of any $f \in \gF$ on any data point $z \in \gD$.  For example, $s(f, z)=\mathbbm{1}[f(x)=y]$ if the benchmark uses standard accuracy as the metric. 
        \item $\bar{s}(f, \gD')=\frac{1}{|\gD'|} \sum_{z \in \gD'} s(f, z)$ represents the average performance of $f \in \gF$ on any $\gD' \subset \gD$.
        \item $\vs(f, \gD')=\{s(f, z)\}_{z \in \gD'}$ represents the vectorized performance of $f \in \gF$ on all data points in $\gD' \subset \gD$, and $\vs(\gF', z)=\{s(f, z)\}_{f \in \gF'}$ represents the vectorized performances of all models in $\gF' \subset \gF$ on data point $z \in \gD$.
        \item $S(\gF', \gD')=\{\vs(f, \gD')\}_{f \in \gF'}=\{\vs(\gF', z)\}^\texttt{T}_{z \in \gD'}$ as the performance matrix of all models in $\gF' \subset \gF$ on all data points in $\gD' \subset \gD$.
    \end{itemize}
    \item  $\gF^{(s)} =\{f_{1}, \ldots, f_{M}\} \subset \gF$ refers to a set of source models, whose performances on every data point of the benchmark $S(\gF^{(s)}, \gD)$ are known. 
    \item The rest of the models are referred to as target models $\gF^{(t)}=\gF \setminus \gF^{(s)}$, which can only be evaluated on at most $n \ll N$ data points to save computational costs.
\end{itemize}

Benchmark prediction with fewer data aims to estimate $\bar{s}(f, \gD)$ for every $f \in \gF^{(t)}$ with only $n$ data points.
In practice, benchmark prediction often involves two steps:  \ding{172} identifying a representative core-set $\gC \subset \gD$ with $|\gC|=n$ data points, and \ding{173} learning a performance estimator $h$ to estimate the average performance on the full benchmark based on the core-set.
Formally, the goal of benchmark prediction  is to find $\gC$ and $h$ to minimize the estimation gap over target models,
\begin{align}
    \text{estimation gap:~~~~~~}
   \frac{1}{|\gF^{(t)}|} \sum_{f \in \gF^{(t)}} \big| \bar{s}(f, \gD) - h\big[ \vs(f, \gC), S(\gF^{(s)}, \gD) \big]  \big|\,\text{.}
\end{align}
For simplicity, in the remainder of the paper, we will denote the estimated performance of target model $f\in\gF^{(t)}$ as $h(f)$, instead of explicitly writing $h[ \vs(f, \gC), S(\gF^{(s)}, \gD)]$.

\subsection{Benchmark Prediction Methods}
\paragraph{Previous methods}
In this paper, we examine five widely-used benchmark prediction methods,
\begin{itemize}[left=4pt]
    \item \textsc{Random-Sampling} randomly samples a subset as $\gC$ and directly returns the mean performance,
    \begin{align}
        h^{\textsc{Random-Sampling}}(f) = \bar{s}(f, \gC)\,\text{.}
    \end{align}
    If the benchmark metric $s$ is standard accuracy, the gap $|\bar{s}(f, \gC) - \bar{s}(f, \gD)|$ is bounded by $\gO(\sqrt{1/n})$ with high probability based on Hoeffding's inequality.
    \item \textsc{Anchor-Points-Weighted}~\citep{anchorpoints} treats benchmark prediction as a k-medoids clustering problem. The selected medoids are used as $\gC$, and a weight vector $\bm{\theta} \in \mathbbm{R}^n$ is calculated as the normalized cluster size of each medoid. 
    The final estimate for any target model $f \in \gF^{(t)}$ is  
    \begin{align}
    h^{\textsc{Anchor-Points-Weighted}}(f) =  \vs(f, \gC)^T \bm{\theta}\,\text{.}
    \end{align}
    \item \textsc{Anchor-Points-Predictor}~\citep{anchorpoints} extends \textsc{Anchor-Points-Weighted}.
    Instead of directly returning the weighted sum, a linear regression model $\vg[\vs(f, \gC)]$ is learned to predict $\vs(f, \gD-\gC)$. 
    \begin{align}
    & h^{\textsc{Anchor-Points-Predictor}}(f) = \bar{g}[\vs(f, \gC)] \\
    \text{~~where~~} & \vg = \argmin_{\vg'} \frac{1}{M} \sum_{f \in \gF^{(s)}} \| \vs(f, \gD-\gC) - \vg'[ \vs(f, \gC) ]  \big\|^2_2\,\text{,} 
    \label{eq:app_loss}
    \end{align}
    where we note that $\vg[\vs(f, \gC)]$ is a $(N-n)$ dimensional vector and we use $\bar{g}[\vs(f, \gC)]$ as its mean.
    \item \textsc{P-IRT}~\citep{tinybench} extends \textsc{Anchor-Points-Predictor} by replacing the regression model $\vg$ in \eqref{eq:app_loss} with an Item Response Theory (IRT) model. Following the notation for \textsc{Anchor-Points-Predictor}, we estimate performance for any $f\in \gF^{(t)}$ as follows:
    \begin{align}
        h^{\textsc{P-IRT}}(f) = \frac{N-n}{N}\bar{g}[\vs(f, \gC)]+\frac{n}{N}\bar{s}(f, \gC)\,\text{.}
    \end{align}
    \item \textsc{GP-IRT}~\citep{tinybench} further generalizes $\textsc{P-IRT}$ by combining its estimation with \textsc{Anchor-Points-Weighted} as a weighted sum,
    \begin{align}
        & h^{\textsc{GP-IRT}}(f) = \lambda h^{\textsc{Anchor-Points-Weighted}}(f) + (1 - \lambda) h^{\textsc{P-IRT}}(f)\,\text{,}
    \end{align}
    where $\lambda$ is chosen heuristically to control the error of \textsc{P-IRT}.
\end{itemize}
\paragraph{New methods}
We introduce six methods that have not yet been applied to benchmark prediction.
\begin{itemize}[left=4pt]
    \item \textsc{Random-Sampling-Learn} randomly samples a subset as $\gC$ and adopts a Ridge regression model $g$ for estimation as follows,
    \begin{align}
        & h^{\textsc{Random-Sampling-Learn}}(f) = g[\vs(f, \gC)] \\
        \text{~~where~~} & g = \argmin_{g'} \frac{1}{M} \sum_{f \in \gF^{(s)}} \big| \bar{s}(f, \gD) - g'[ \vs(f, \gC) ]  \big|\,\text{.}
    \end{align}
    \item \textsc{Random-Search-Learn} performs \textsc{Random-Sampling-Learn} for 10,000 times and selects the best-performing subset as $\gC$ based on cross-validation. A Ridge regression model $g$ is then trained and used in the same way as \textsc{Random-Selection-Learn}.
    \item \textsc{Lasso} trains a Lasso regression model with weights $\bm{\theta} \in \mathbbm{R}^N$ as follows, 
    \begin{align}
        & h^{\textsc{Lasso}}(f) = \vs(f, \gC)^T\bm{\theta}_{\gC} \\
        \text{where~~} 
        \bm{\theta} = \argmin_{\bm{\theta}'}& \frac{1}{n} \sum_{z \in \gC} \big[
        \vs(f, \gD)^{\texttt{T}} \bm{\theta}' - \bar{s}(f, \gD)
        \big]^2 + \lambda \|\bm{\theta}'\|_1\,\text{,}
    \end{align}
    where $\lambda$ is selected so that only $n$ dimensions of $\bm{\theta}$ are non-zero and $\bm{\theta}_{\gC}$ is the non-zero slice of $\bm{\theta}$.
    \item \textsc{Double-Optimize} optimizes both a subset selection vector $\bm{\pi} \in \mathbbm{R}^N$ and a linear regression model with weights $\bm{\theta} \in \mathbbm{R}^N$ with gradient descent as follows,
    \begin{align}
    & h^{\textsc{Double-Optimize}}(f) = [\vs(f, \gD)\cdot \texttt{TopMask}(\bm{\pi}; n) ] ^{\texttt{T}} \bm{\theta} \\
    \text{where~~~~} &
    \bm{\pi}, \bm{\theta} =  \argmin_{\bm{\pi}',\bm{\theta}'}
        \big\{
        [\vs(f, \gD) \cdot \texttt{TopMask}(\bm{\pi}'; n)]^{\texttt{T}} \bm{\theta}' - \bar{s}(f, \gD) 
        \big\} ^2\,\text{,}
    \end{align}
    where $\cdot$ refers to the bitwise multiplication between two vectors, and $\texttt{TopMask}(\bm{\pi}'; n)$ replaces the top $n$ largest values of $\bm{\pi}'$ with 1s and the rest with 0s.
    We directly pass the gradient on $\texttt{TopMask}(\bm{\pi}'; n)$ to $\bm{\pi}'$ during optimization following the Straight-Through technique~\cite{Jang2016CategoricalRW,Bengio2013EstimatingOP}.
    \item Principal Component Analysis (\textsc{PCA}) treats benchmark prediction as a matrix completion problem. This method assumes the performance matrix $S(\gF, \gD)$ is of low rank. 
    By randomly sampling a subset as $\gC$, this methods conducts PCA to impute the missing values for target models~\citep{Vershynin2016FourLO,cai2010singular}.
    As a more intuitive view, one could also take the acquired principal components as model capability indicators~\citep{Ruan2024ObservationalSL}, \emph{i.e.}, the $(M \times k)$ PCA-transformed scores indicate the $k$-capabilities of each model, while the $(k \times N)$ principal components represent the capability requirements for each data point. 
    We select $k$ among $\{2, 5, 10, 20\}$ through cross-validation.
    The Pseudo codes are in Algorithm~\ref{alg:pca}.
    \item Augmented inverse propensity weighting (AIPW) \cite{robins1995semiparametric}: Inspired by the application of prediction powered inference ~\citep{Angelopoulos2023PPIEP,Angelopoulos2023PredictionpoweredI} to the LLM-as-a-judge setting \cite{boyeau2024autoeval,Dorner2024LimitsTS}, we apply a more general AIPW estimator to benchmark prediction. 
    We train a Ridge regression model $g$ for every target model $f$, which predicts the point-wise performance $s(f,z)$ based on $\vs(\gF^{(s)}, z)$.
    Formally, 
    \begin{align}
        g = \argmin_{g'} \frac{1}{n} \sum_{z \in \gC} \big[
        g'[\vs(\gF^{(s)}, z)] - s(f, z)
        \big]^2 \text{.}
    \end{align}
    The idea behind the AIPW estimator is to use the predicted performance $\hat{s}(f, z)=g[\vs(\gF^{(s)}, z)]$ as a proxy score to estimate $\bar{s}(f,\mathcal{D})$ and ''debias'' that estimator as follows
    \begin{align}
    h^{\textsc{AIPW}}(f) = \bar{s}(f, \gC) + \frac{1}{1+\frac{n}{N-n}}\left(\frac{1}{N-n} \sum_{z \in \gD-\gC} \hat{s}(f, z)\ - \frac{1}{n} \sum_{z \in \gC} \hat{s}(f, z)\right) \text{.}
    \end{align}
    Unlike the other learning-based baselines, AIPW is a consistent estimator for $\bar{s}(f, \gD)$\cite{glynn2010introduction}. Compared to \textsc{Random-Sampling}, it reduces estimator variance by a factor of up to $\frac{1}{1+\frac{n}{N}} \rho(\hat{s}^(f,z),s(f,z))^2$ \cite{Dorner2024LimitsTS}, where $\rho$ is the Pearson correlation coefficient.
    Recent research~\citep{Mani2025NoFL} shows that AIPW estimator will outperform random sampling if and only if the correlation between $\hat{s}(f,z)$ and $s(f,z)$ is above a certain level that depends on $n$.
\end{itemize}

\begin{algorithm}[H]
\caption{PCA Impute Process}
\label{alg:pca}
    \begin{algorithmic}[1]
    \STATE \textbf{Input:} Data matrix with missing values
    \STATE \textbf{Parameters:} number of components $k$, max iteration \texttt{max\_iter}, stopping threshold \texttt{tol}
    \STATE \textbf{Output:} Imputed data matrix
    \STATE \textbf{Step 1: Initialization}
    \STATE \quad Compute initial values for missing entries using column means
    \STATE \textbf{Step 2: Iterative Imputation}
    \FOR{iteration $\leftarrow 1$ to \texttt{max\_iter}}
        \STATE \textbf{PCA Decomposition:}
        \STATE \quad Perform PCA retaining $k$ components
        \STATE \quad Transform data to the lower-dimensional space
        \STATE \quad Reconstruct the data from the lower-dimensional space
    
        \STATE \textbf{Evaluate Convergence:}
        \STATE \quad Compute the norm of differences between imputed and original values at missing entries
        \IF{norm $<$ \texttt{tol}}
            \STATE Break the loop
        \ENDIF
    
        \STATE \textbf{Update Imputed Values:}
        \STATE \quad Replace missing values with reconstructed values
    \ENDFOR
    \STATE \RETURN Fully imputed data matrix
    \end{algorithmic}
\end{algorithm}

\newpage
\section{Additional Experiment Setup}
\label{app:setup}
We select a diverse range of benchmarks from the following sources\footnote{Since \textsc{P-IRT} and \textsc{GP-IRT} requires $s(f, z)$ to be binary, we only use benchmarks with accuracy as metric.}.
\begin{itemize}[left=0pt]
    \item {HELM-Lite} benchmarks~\citep{Liang2023HolisticEO}:
    \begin{itemize}
    \item \texttt{OpenbookQA}~\citep{openbookqa}: $N=500$ data points.
    \item \texttt{GSM8K}~\citep{gsm8k}: $N=1000$ data points.
    \item \texttt{LegalBench}~\citep{legalbench}: $N=2047$ data points.
    \item \texttt{Math}~\citep{math}: $N=437$ data points.
    \item \texttt{MedQA}~\citep{MedQA}: $N=1000$ data points.
    \item \texttt{MMLU}~\citep{mmlu}: $N=567$ data points. 
    \end{itemize}
    We obtain the per-data point performances of $|\gF|=83$ models from the official leaderboard. 
    Note that Helm-Lite often only uses a subset of the original testing set for each benchmark to save compute.
    \item {GLUE} benchmarks~\citep{Wang2018GLUEAM}:
    \begin{itemize}
        \item \texttt{MRPC}~\citep{mrpc}: $N=408$ data points.
        \item \texttt{RTE}~\citep{rte,rte2,rte3}: $N=277$ data points.
        \item \texttt{SST-2}~\citep{sst2}: $N=872$ data points.
        \item \texttt{MNLI}~\citep{mnli}: $N=9815$ data points.
        \item \texttt{QNLI}~\citep{qnli}:  $N=5463$ data points.
    \end{itemize}
    We use the per-data performances of $|\gF|=87$ models provided by AnchorPoint\footnote{The provided score file for \texttt{QQP} is broken so we exclude it.}~\citep{anchorpoints}.
    \item {OpenLLM} benchmarks~\citep{open-llm-leaderboard-v2}: 
    \begin{itemize}
    \item \texttt{IFEval}~\citep{ifeval}: $N=541$ data points.
    \item \texttt{Math}~\citep{math}: $N=894$ data points. Only level 5 MATH questions are used in OpenLLM.
    \item \texttt{MMLU-Pro}~\citep{mmlupro}: $N=12032$ data points.
    \item \texttt{Arc-Challenge}~\citep{allenai:arc}: $N=1172$ data points.
    \item \texttt{BBH}~\citep{bbh}: $N=5761$ data points.
    \item \texttt{GPQA}~\citep{gpqa}: $N=1192$ data points.
    \item \texttt{MUSR}~\citep{musr}: $N=756$ data points.
    \end{itemize}
    We use $|\gF|=448$ models provided by Huggingface~\footnote{\url{https://huggingface.co/spaces/open-llm-leaderboard/open_llm_leaderboard\#}} and collect their performance scores. 
    \item \texttt{ImageNet}~\citep{imagenet}: We collect $|\gF|=110$ models from Pytorch Hub~\footnote{\url{https://pytorch.org/vision/stable/models.html\#classification}} and evaluate them on \texttt{ImageNet} with $N=50,000$ data points.
\end{itemize}
For simplicity, we report the overall average accuracy directly for \texttt{MMLU}, \texttt{MMLU-Pro}, and \texttt{BBH}, rather than the weighted average accuracy computed across sub-tasks.
Alternatively, one could apply benchmark predictions separately to each sub-task and then calculate the weighted average accuracy.

\section{Additinoal Experiment Results}
\label{app:results}

\begin{table}[]
    \caption{
    Training and inference time of each method on \texttt{ImageNet} with $N=50000$ data points and $|\gF|=110$ models. Training is based on 83 source models, and inference is on 27 target models.
    }
    \label{app_tab:time}
    \centering
    \resizebox{0.7\textwidth}{!}{
    \begin{tabular}{ccc}
    \toprule
    \midrule
     & Training Time (s) & Inference Time (s) \\
    \midrule
    \textsc{Random-Sampling} & 0.00 & 0.00 \\
    \textsc{Random-Sampling-Learn} & 0.02 & 0.00 \\
    \textsc{PCA} & 0.59 & 19.20 \\
    \textsc{AIPW} & 0.00 & 0.27 \\
    \textsc{Random-Search-Learn} & 81.02 & 0.00 \\
    \textsc{Lasso} & 105.58 & 0.01 \\
    \textsc{Double-Optimize} & 4.88 & 0.00 \\
    \textsc{Anchor-Points-Weighted} & 84.26 & 0.00 \\
    \textsc{Anchor-Points-Predictor} & 197.71 & 0.26 \\
    \textsc{P-IRT} & 585.72 & 0.90 \\
    \textsc{GP-IRT} & 1750.20 & 0.89 \\
    \midrule
    \bottomrule
    \end{tabular}}
\end{table}

\begin{table}[t]
\caption{
Average estimation gap between the predicted rankings based on the coreset and the actual rankings based on the full benchmark, measured by Kendall's $\tau$ ($\uparrow$).
The results are averaged over all benchmarks.
}
\label{tab:ranking}
\centering
\resizebox{\textwidth}{!}{
\begin{tabular}{c|ccccc|ccccc}
\toprule
\midrule
&  \multicolumn{5}{c|}{Interpolation} & \multicolumn{5}{c}{Extrapolation} \\
& {$n=10$}
& {$n=20$}
& {$n=50$}
& {$n=100$}
& {$n=200$}
& {$n=10$}
& {$n=20$}
& {$n=50$}
& {$n=100$}
& {$n=200$}\\
\midrule
\midrule
\textsc{Random-Sampling} & 0.52 & 0.61 & 0.70 & 0.78 & 0.84 & 0.36 & 0.43 & 0.53 & 0.63 & 0.73 \\
\textsc{Random-Sampling-Learn} & 0.57 & 0.66 & 0.75 & 0.81 & 0.86 & 0.07 & 0.12 & 0.18 & 0.27 & 0.36 \\
\textsc{PCA} & 0.55 & 0.63 & 0.72 & 0.78 & 0.83 & 0.04 & 0.10 & 0.21 & 0.40 & 0.57 \\
\textsc{AIPW} & 0.52 & 0.62 & 0.72 & 0.79 & 0.85 & 0.31 & 0.39 & 0.51 & 0.61 & 0.71 \\
\textsc{Random-Search-Learn} & 0.66 & 0.70 & 0.76 & 0.82 & 0.86 & 0.13 & 0.13 & 0.20 & 0.29 & 0.38 \\
\textsc{Lasso} & 0.68 & 0.71 & 0.77 & 0.81 & 0.82 & 0.05 & 0.06 & 0.12 & 0.19 & 0.22 \\
\textsc{Double-Optimize} & 0.58 & 0.66 & 0.76 & 0.81 & 0.84 & 0.31 & 0.36 & 0.44 & 0.50 & 0.58 \\
\textsc{Anchor-Points-Weighted} & 0.65 & 0.70 & 0.76 & 0.81 & 0.85 & 0.37 & 0.43 & 0.50 & 0.60 & 0.69 \\
\textsc{Anchor-Points-Predictor} & 0.67 & 0.72 & 0.77 & 0.80 & 0.80 & 0.21 & 0.25 & 0.32 & 0.38 & 0.44 \\
\textsc{P-IRT} & 0.52 & 0.58 & 0.71 & 0.80 & 0.87 & 0.28 & 0.31 & 0.42 & 0.56 & 0.69 \\
\textsc{GP-IRT} & 0.53 & 0.59 & 0.72 & 0.80 & 0.86 & 0.28 & 0.33 & 0.45 & 0.59 & 0.71 \\
\midrule
\bottomrule
\end{tabular}}
\end{table}

\newpage
\subsection{Detailed Results}
\begin{figure}
    \centering
    \includegraphics[width=0.99\linewidth]{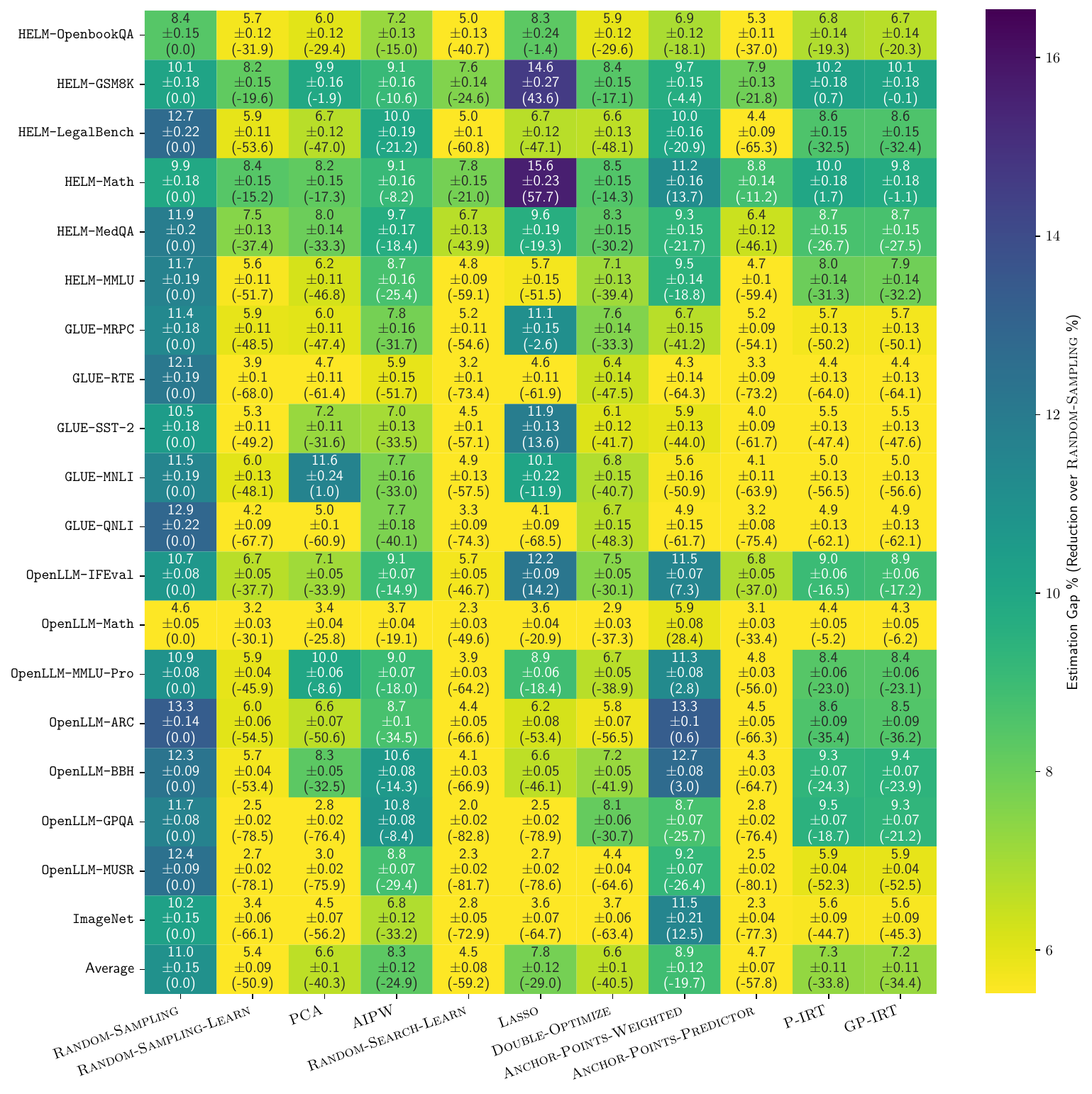}
    \caption{
    The estimation gaps ($\downarrow$) for target models (calculated as \eqref{eq:problem}) under interpolation model split, where source models are identically distributed with target models.
    Each target model can only be evaluated on $n=10$ data points.
    We also report $\pm$ the standard error of the mean and the estimation gap reduction ($\downarrow$) over \textsc{Random-Sampling} in parentheses. 
    A negative reduction implies that the method achieves a lower estimation gap than \textsc{Random-Sampling}.
    \% is omitted.
    Best viewed in color.
     }
    \label{app_fig:acc_gap_10}
\end{figure}

\begin{figure}
    \centering
    \includegraphics[width=0.99\linewidth]{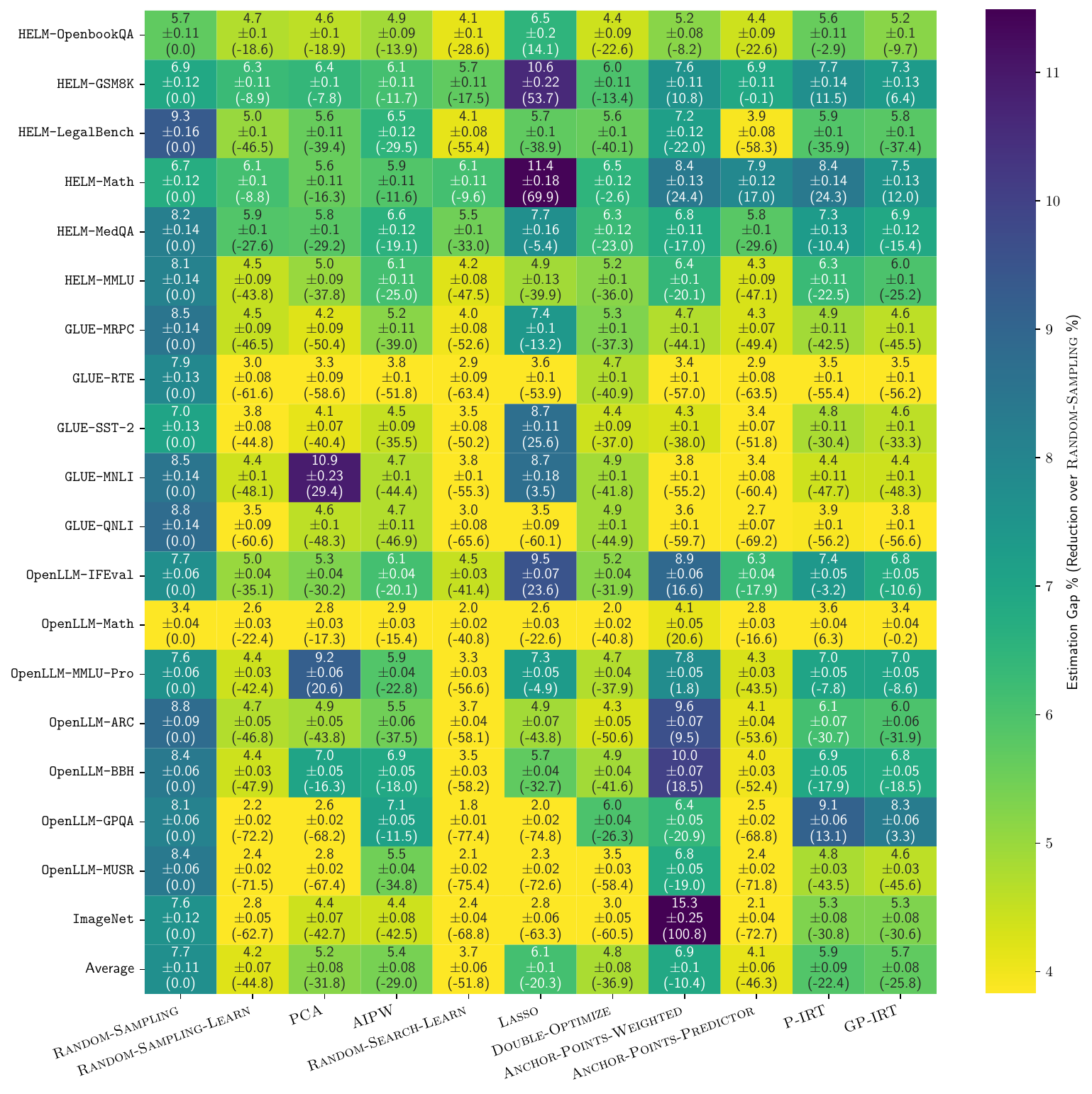}
    \caption{
    The estimation gaps ($\downarrow$) for target models (calculated as \eqref{eq:problem}) under interpolation model split, where source models are identically distributed with target models.
    Each target model can only be evaluated on $n=20$ data points.
    We also report $\pm$ the standard error of the mean and the estimation gap reduction ($\downarrow$) over \textsc{Random-Sampling} in parentheses. 
    A negative reduction implies that the method achieves a lower estimation gap than \textsc{Random-Sampling}.
    \% is omitted.
    Best viewed in color.
     }
    \label{app_fig:acc_gap_20}
\end{figure}

\begin{figure}
    \centering
    \includegraphics[width=0.99\linewidth]{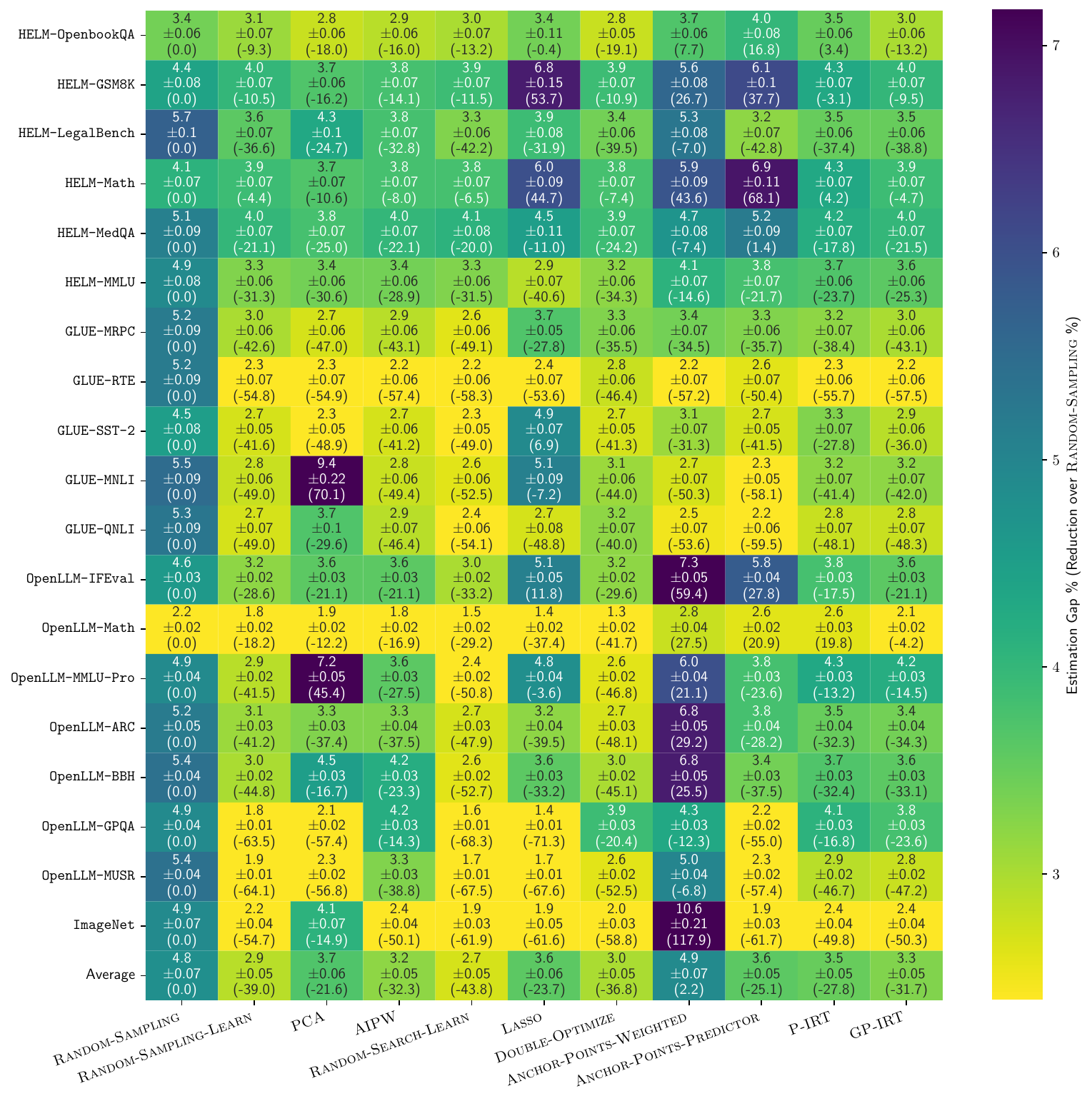}
    \caption{
    The estimation gaps ($\downarrow$) for target models (calculated as \eqref{eq:problem}) under interpolation model split, where source models are identically distributed with target models.
    Each target model can only be evaluated on $n=50$ data points.
    We also report $\pm$ the standard error of the mean and the estimation gap reduction ($\downarrow$) over \textsc{Random-Sampling} in parentheses. 
    A negative reduction implies that the method achieves a lower estimation gap than \textsc{Random-Sampling}.
    \% is omitted.
    Best viewed in color.
     }
    \label{app_fig:acc_gap}
\end{figure}

\begin{figure}
    \centering
    \includegraphics[width=0.99\linewidth]{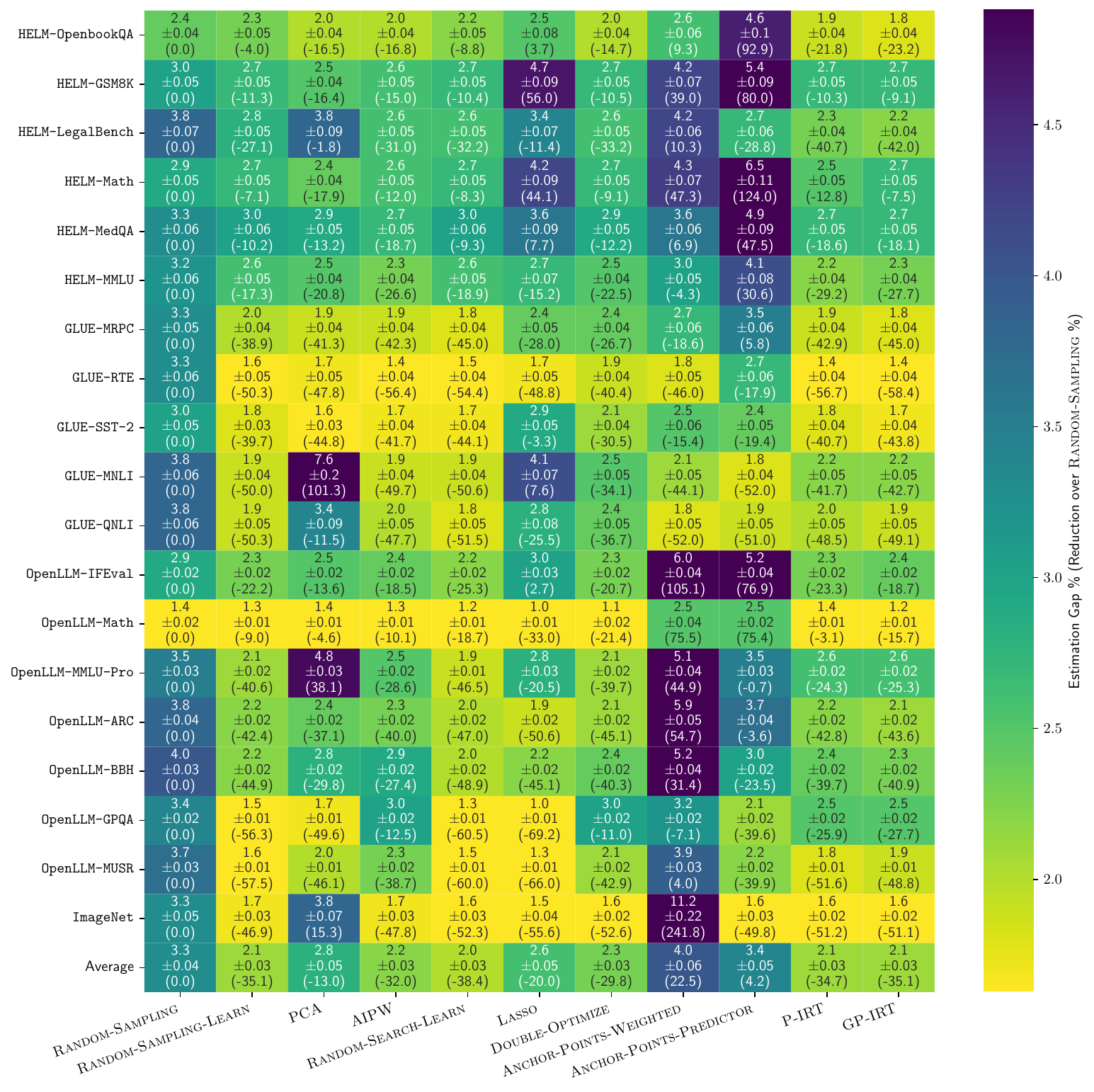}
    \caption{
    The estimation gaps ($\downarrow$) for target models (calculated as \eqref{eq:problem}) under interpolation model split, where source models are identically distributed with target models.
    Each target model can only be evaluated on $n=100$ data points.
    We also report $\pm$ the standard error of the mean and the estimation gap reduction ($\downarrow$) over \textsc{Random-Sampling} in parentheses. 
    A negative reduction implies that the method achieves a lower estimation gap than \textsc{Random-Sampling}.
    \% is omitted.
    Best viewed in color.
     }
    \label{app_fig:acc_gap_100}
\end{figure}

\begin{figure}
    \centering
    \includegraphics[width=0.99\linewidth]{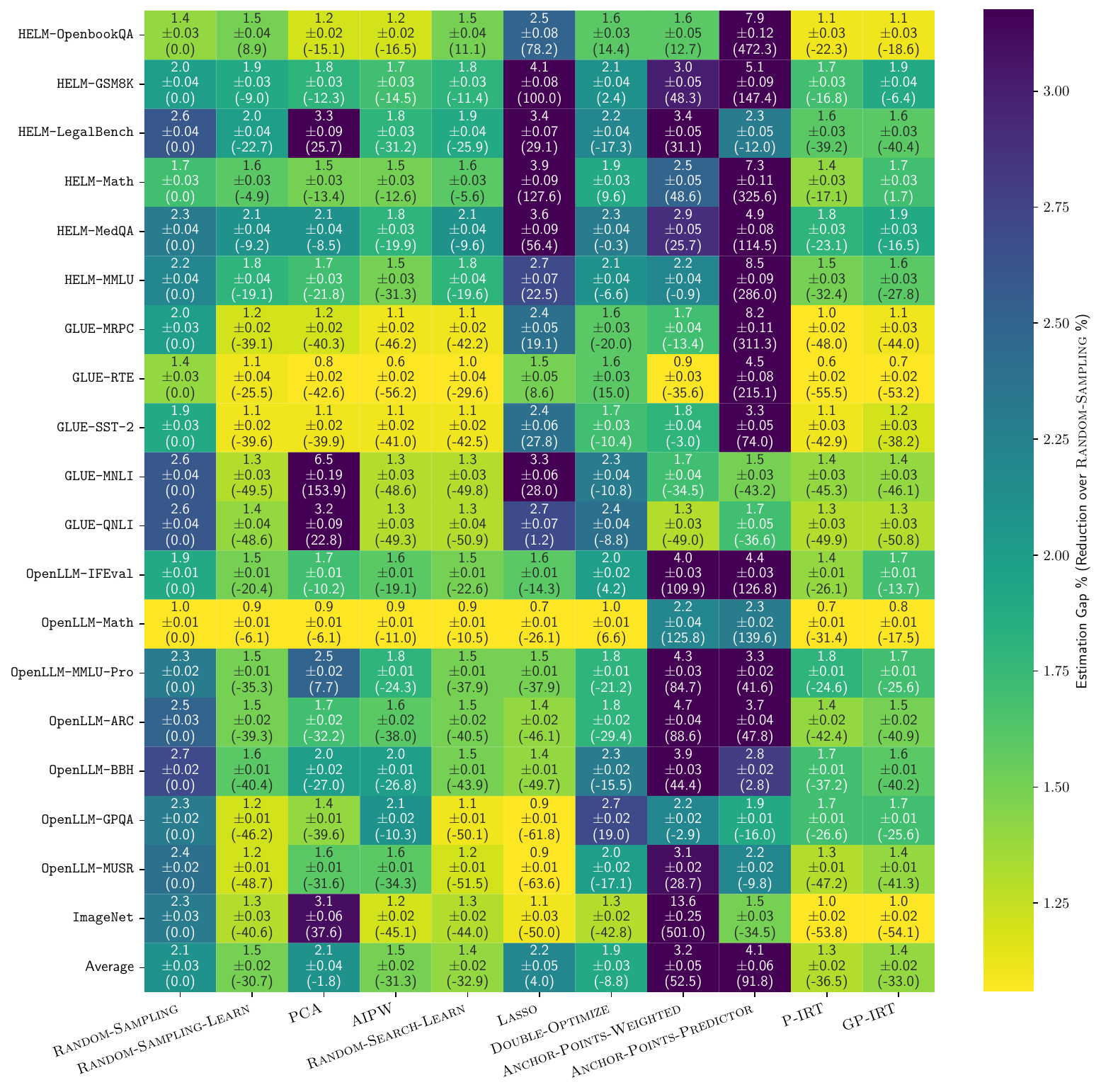}
    \caption{
    The estimation gaps ($\downarrow$) for target models (calculated as \eqref{eq:problem}) under interpolation model split, where source models are identically distributed with target models.
    Each target model can only be evaluated on $n=200$ data points.
    We also report $\pm$ the standard error of the mean and the estimation gap reduction ($\downarrow$) over \textsc{Random-Sampling} in parentheses. 
    A negative reduction implies that the method achieves a lower estimation gap than \textsc{Random-Sampling}.
    \% is omitted.
    Best viewed in color.
     }
    \label{app_fig:acc_gap_200}
\end{figure}

\begin{figure}
    \centering
    \includegraphics[width=0.99\linewidth]{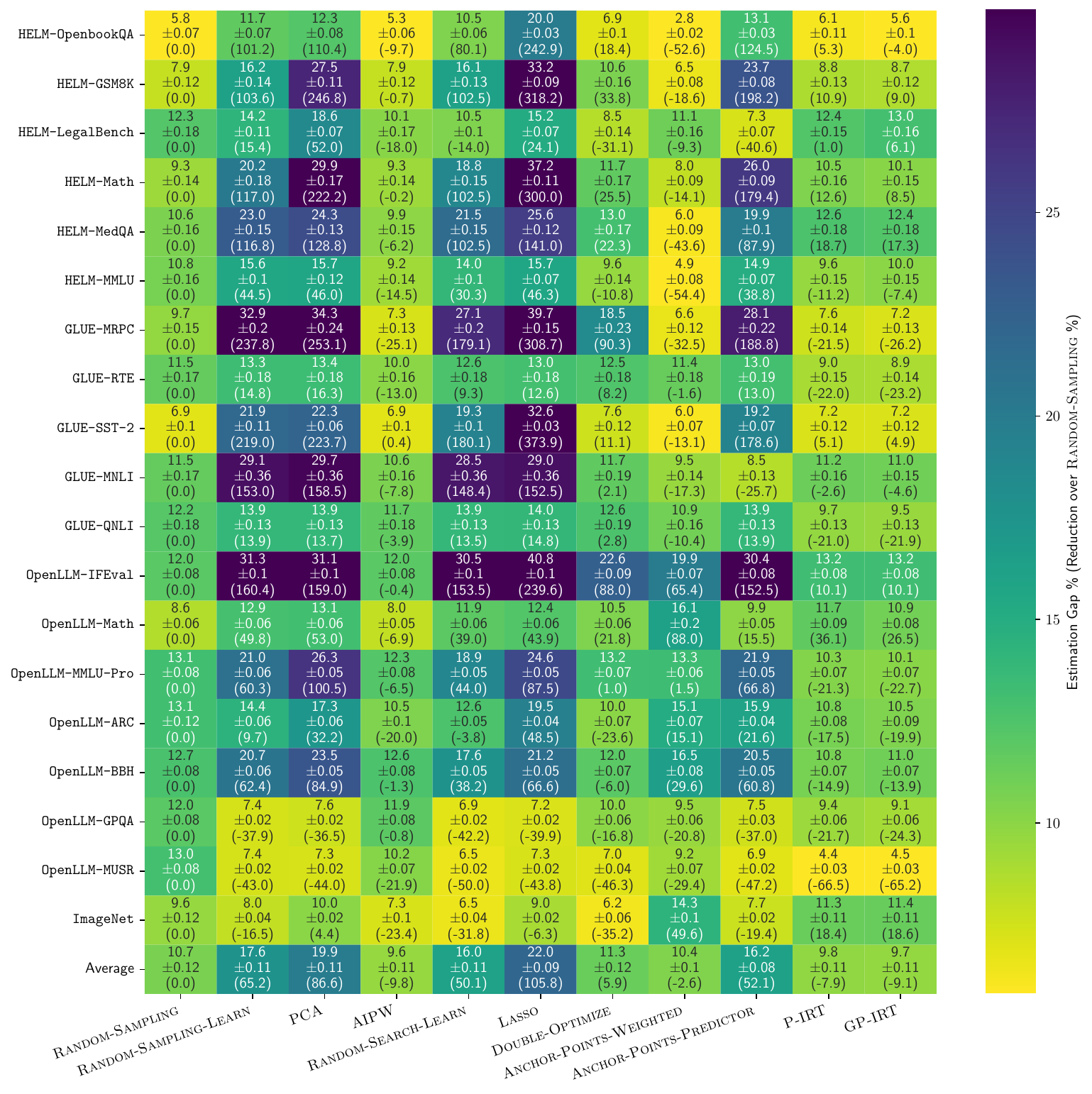}
    \caption{
    The estimation gaps ($\downarrow$) for target models (calculated as \eqref{eq:problem}) under extrapolation model split, where source models are the lowest-performing 50\%, and target models are the top 30\% based on average performance over the full benchmark.
    Each target model can only be evaluated on $n=10$ data points.
    We also report $\pm$ the standard error of the mean and the estimation gap reduction ($\downarrow$) over \textsc{Random-Sampling} in parentheses. 
    A negative reduction implies that the method achieves a lower estimation gap than \textsc{Random-Sampling}.
    \% is omitted.
    Best viewed in color.
     }
    \label{app_fig:acc_gap_top_10}
\end{figure}

\begin{figure}
    \centering
    \includegraphics[width=0.99\linewidth]{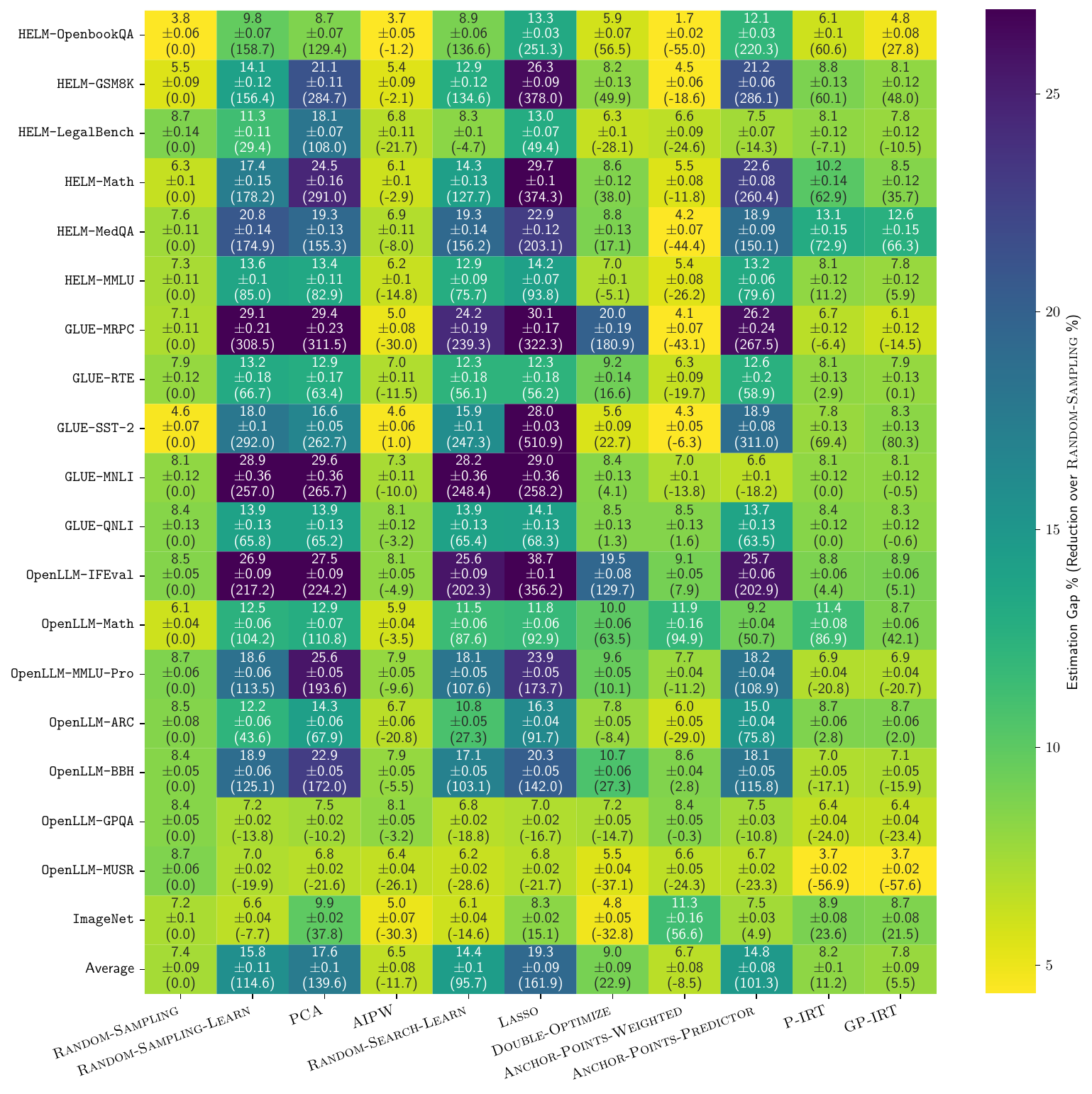}
    \caption{
    The estimation gaps ($\downarrow$) for target models (calculated as \eqref{eq:problem}) under extrapolation model split, where source models are the lowest-performing 50\%, and target models are the top 30\% based on average performance over the full benchmark.
    Each target model can only be evaluated on $n=20$ data points.
    We also report $\pm$ the standard error of the mean and the estimation gap reduction ($\downarrow$) over \textsc{Random-Sampling} in parentheses. 
    A negative reduction implies that the method achieves a lower estimation gap than \textsc{Random-Sampling}.
    \% is omitted.
    Best viewed in color.
     }
    \label{app_fig:acc_gap_top_20}
\end{figure}

\begin{figure}
    \centering
    \includegraphics[width=0.99\linewidth]{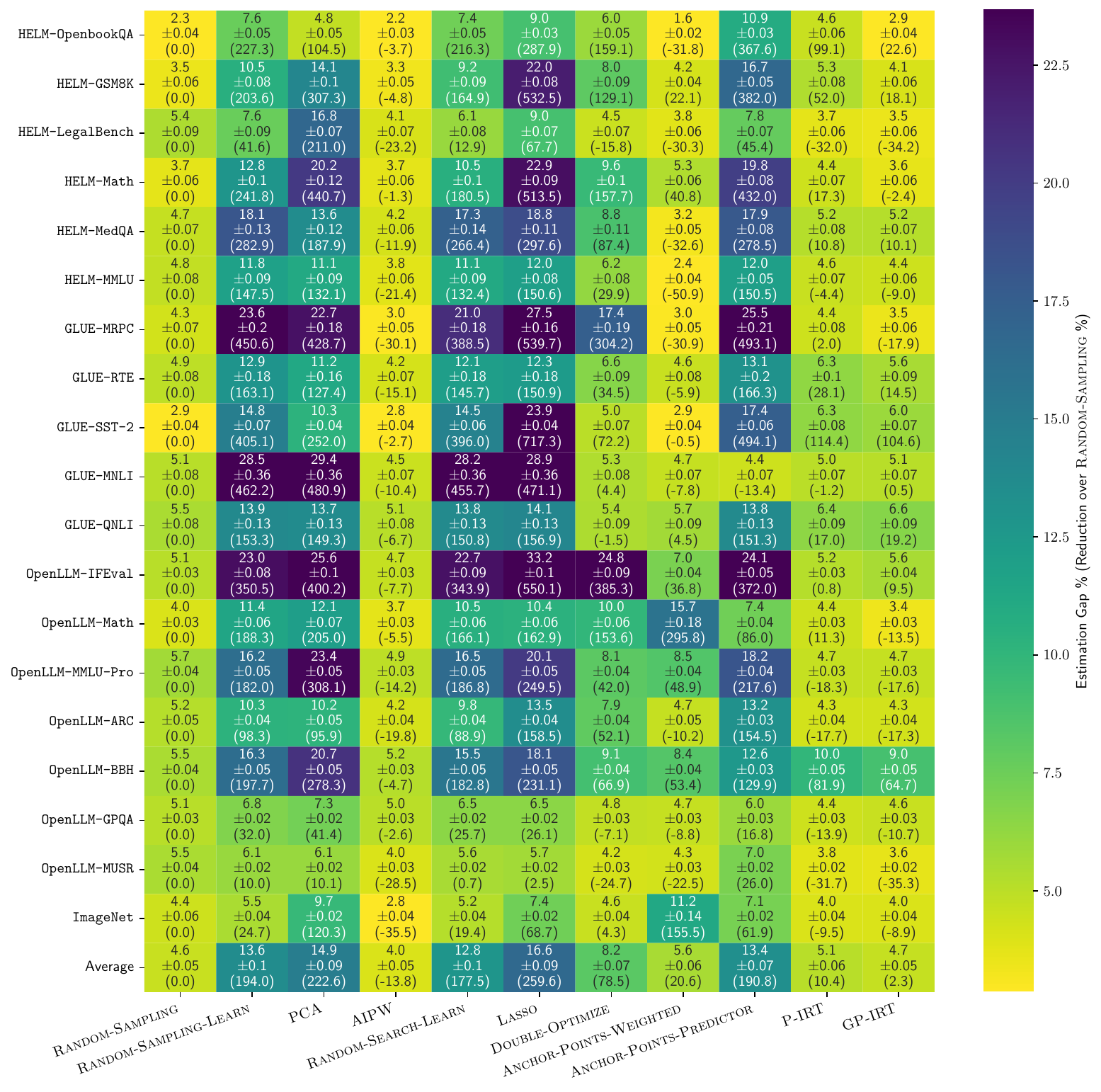}
    \caption{
    The estimation gaps ($\downarrow$) for target models (calculated as \eqref{eq:problem}) under extrapolation model split, where source models are the lowest-performing 50\%, and target models are the top 30\% based on average performance over the full benchmark.
    Each target model can only be evaluated on $n=50$ data points.
    We also report $\pm$ the standard error of the mean and the estimation gap reduction ($\downarrow$) over \textsc{Random-Sampling} in parentheses. 
    A negative reduction implies that the method achieves a lower estimation gap than \textsc{Random-Sampling}.
    \% is omitted.
    Best viewed in color.
     }
    \label{app_fig:acc_gap_top}
\end{figure}

\begin{figure}
    \centering
    \includegraphics[width=0.99\linewidth]{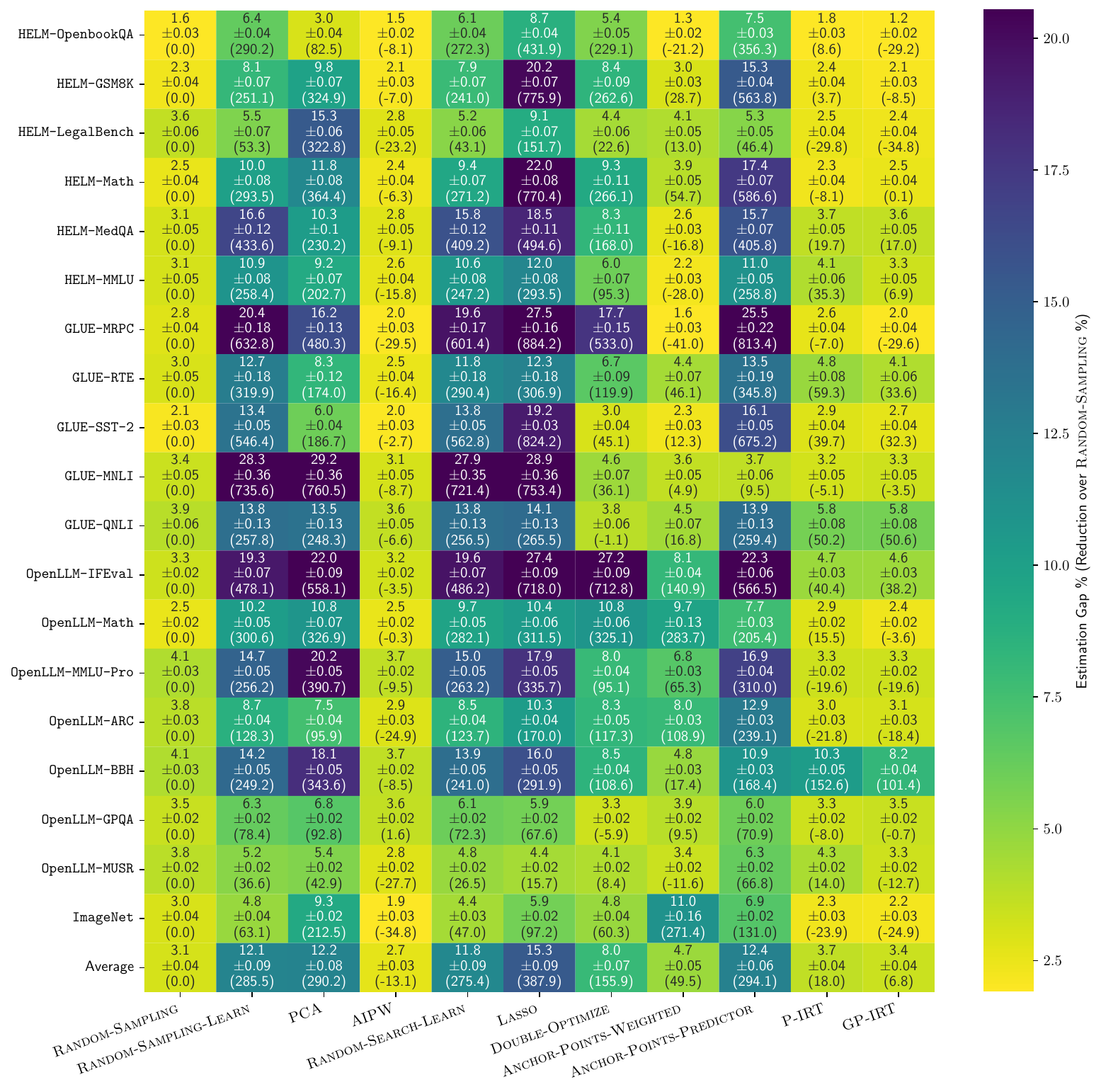}
    \caption{
    The estimation gaps ($\downarrow$) for target models (calculated as \eqref{eq:problem}) under extrapolation model split, where source models are the lowest-performing 50\%, and target models are the top 30\% based on average performance over the full benchmark.
    Each target model can only be evaluated on $n=100$ data points.
    We also report $\pm$ the standard error of the mean and the estimation gap reduction ($\downarrow$) over \textsc{Random-Sampling} in parentheses. 
    A negative reduction implies that the method achieves a lower estimation gap than \textsc{Random-Sampling}.
    \% is omitted.
    Best viewed in color.
     }
    \label{app_fig:acc_gap_top_100}
\end{figure}

\begin{figure}
    \centering
    \includegraphics[width=0.99\linewidth]{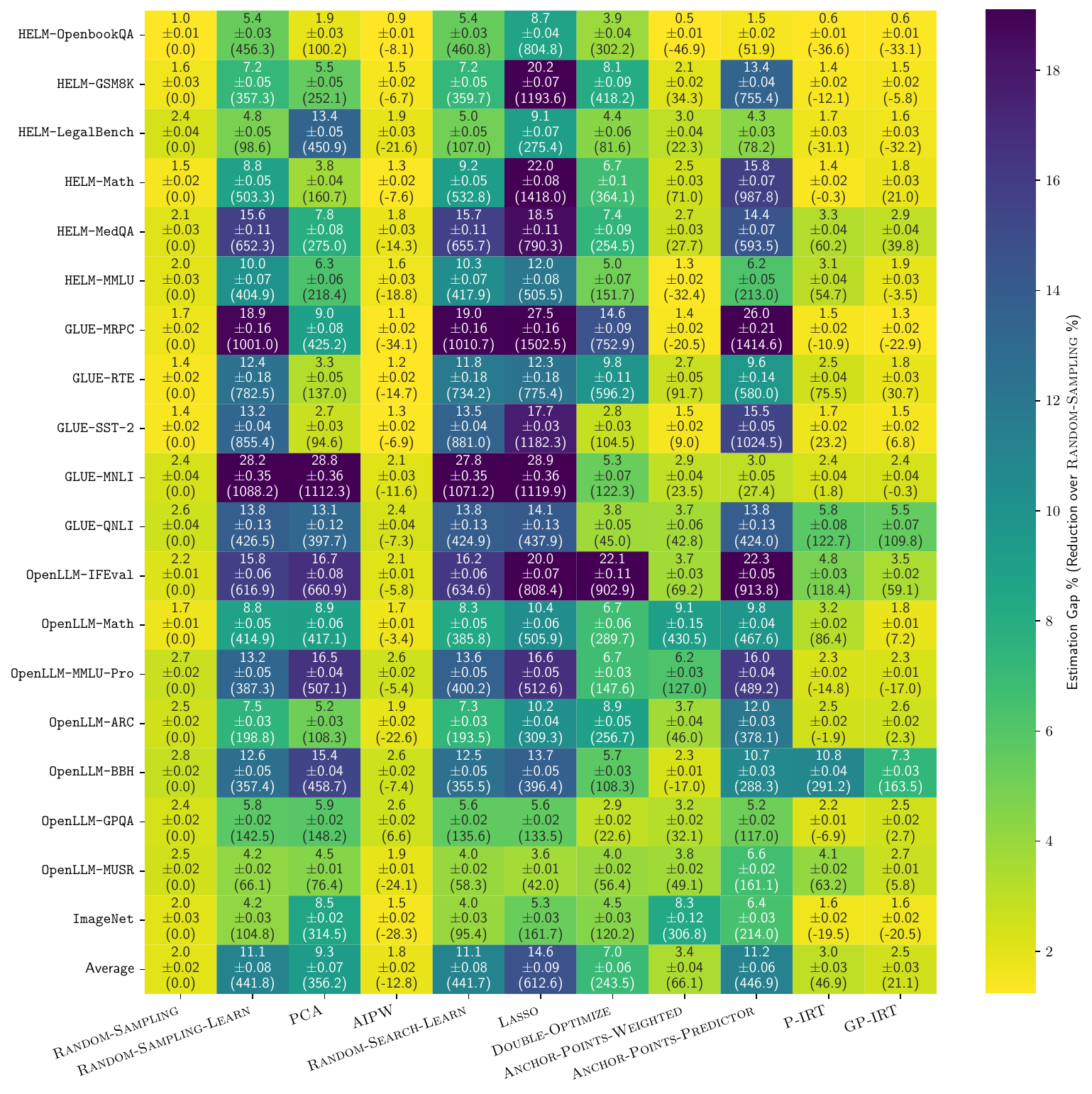}
    \caption{
    The estimation gaps ($\downarrow$) for target models (calculated as \eqref{eq:problem}) under extrapolation model split, where source models are the lowest-performing 50\%, and target models are the top 30\% based on average performance over the full benchmark.
    Each target model can only be evaluated on $n=200$ data points.
    We also report $\pm$ the standard error of the mean and the estimation gap reduction ($\downarrow$) over \textsc{Random-Sampling} in parentheses. 
    A negative reduction implies that the method achieves a lower estimation gap than \textsc{Random-Sampling}.
    \% is omitted.
    Best viewed in color.
     }
    \label{app_fig:acc_gap_top_200}
\end{figure}

In this paper, we experiment with $n \in \{10, 20, 50, 100, 200\}$ under both interpolation and extrapolation settings. 
The detailed results,  along with standard errors,  are reported in  Figures~\ref{app_fig:acc_gap_10}, \ref{app_fig:acc_gap_20}, \ref{app_fig:acc_gap}, \ref{app_fig:acc_gap_100}, \ref{app_fig:acc_gap_200}, \ref{app_fig:acc_gap_top_10}, \ref{app_fig:acc_gap_top_20}, \ref{app_fig:acc_gap_top}, \ref{app_fig:acc_gap_top_100}, and \ref{app_fig:acc_gap_top_200}. 

\subsection{Running time}
While some of the benchmark prediction methods could potentially benefit from the use of GPUs, we opted to run all methods without them, as they are sufficiently fast on standard hardware. 
Table~\ref{app_tab:time} presents the training and inference times for each method on \texttt{ImageNet}. 
Among the models, \textsc{GP-IRT} is the slowest during training because it involves fitting a large Item Response Theory (IRT) model. 
During inference, \textsc{PCA} is the slowest, as it requires multiple imputations of the entire matrix. 
Although \textsc{AIPW} needs training a separate regressor for each target model during inference, the regressor is small, making the inference process remain efficient.

\subsection{Ranking Preservation}
We further compare the predicted rankings of target models with the actual rankings based on the full benchmark using Kendall's $\tau$.
Specifically, we calculate Kendall's $\tau$ for each random trial and average the results over 100 trials.
Our conclusions mostly remain unchanged, with almost all benchmark prediction methods outperforming Random Sampling under interpolation, while none can surpass \textsc{Random-Sampling} under extrapolation. 

\subsection{Case Studies}
We further investigate two additional experimental settings that deviate from the primary setting in the main paper.

\begin{figure}[t]
    \centering
    \includegraphics[width=0.99\linewidth]{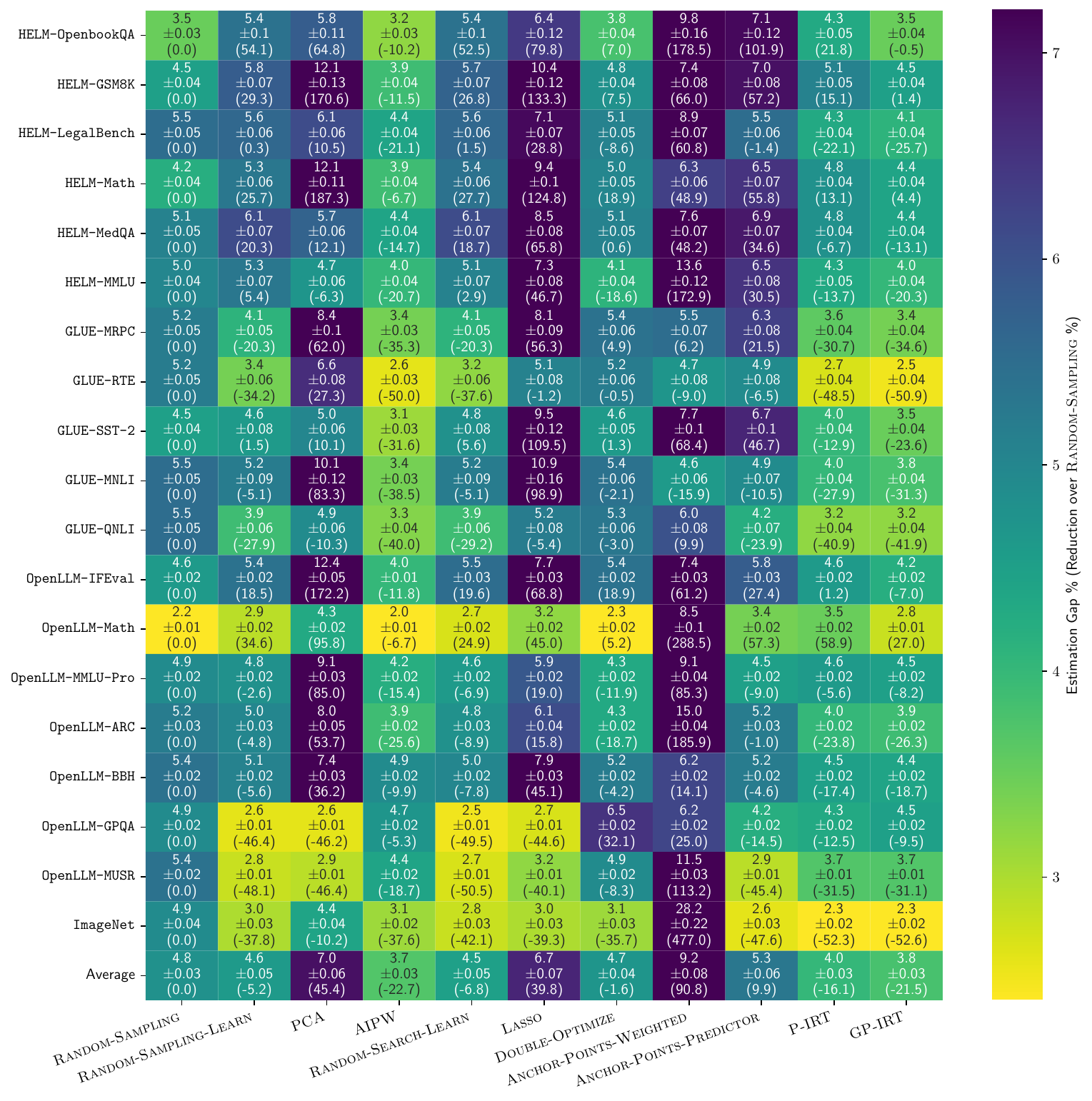}
    \caption{
    Experiment with fewer source models (randomly selected 10 models as source models) under the interpolation model split.
    We report the estimation gaps ($\downarrow$) for target models (calculated as \eqref{eq:problem}).
    We also report $\pm$ the standard error of the mean and the estimation gap reduction ($\downarrow$) over \textsc{Random-Sampling} in parentheses. 
    A negative reduction implies that the method achieves a lower estimation gap than \textsc{Random-Sampling}.
    \% is omitted.
    Best viewed in color.
     }
    \label{app_fig:acc_gap_10_source_50}
\end{figure}

\paragraph{Fewer source models under interpolation.}
Different from the previous interpolation setting that utilized 75\% of models as source models, we now use only 10 models as source models for each benchmark and use the rest as target models. 
All other settings remain unchanged.
This setting allows us to assess the effectiveness of benchmark prediction when ``training data'' from source models is more limited.
Results are shown in Figure~\ref{app_fig:acc_gap_10_source_50}.
Consistent with the findings in the paper, most methods still outperform \textsc{Random-Sampling}, while \textsc{AIPW} and \textsc{GP-IRT} are the best-performing methods.

\begin{figure}[t]
    \centering
    \includegraphics[width=0.99\linewidth]{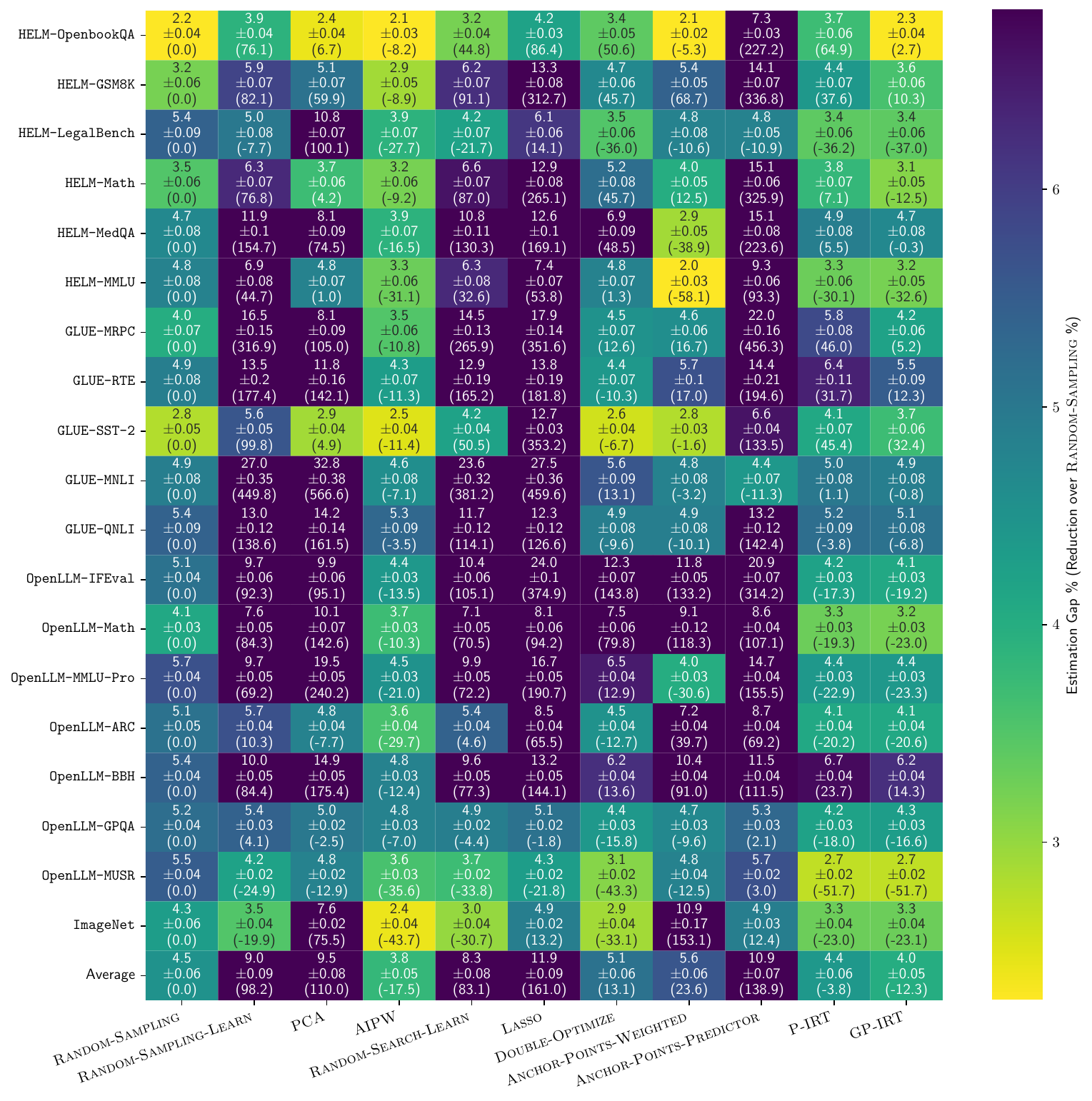}
    \caption{
    Experiment with the near extrapolation model split by using the top 25\% of available models as target models and the remaining bottom 75\% models as source models.
    We report the estimation gaps ($\downarrow$) for target models (calculated as \eqref{eq:problem}).
    We also report $\pm$ the standard error of the mean and the estimation gap reduction ($\downarrow$) over \textsc{Random-Sampling} in parentheses. 
    A negative reduction implies that the method achieves a lower estimation gap than \textsc{Random-Sampling}.
    \% is omitted.
    Best viewed in color.
     }
    \label{app_fig:acc_gap_near_extrapolation}
\end{figure}

\paragraph{Near extrapolation.} 
We modify the previous extrapolation setting, which used the lowest-performing 50\% of models as source models and the top 30\% as target models. In this new setting, we designate the top 25\% of models as target models and utilize all remaining models as source models. 
All other settings remain unchanged.
This setup enables us to examine whether benchmark prediction methods demonstrate improved performance when the distribution gap between source and target models is reduced.
Results are shown in Figure~\ref{app_fig:acc_gap_near_extrapolation}.
Consistent with the findings in the paper, most methods fail to consistently outperform \textsc{Random-Sampling}, except for \textsc{AIPW}. 

\clearpage
\newpage
\section{Broarder Impacts and Limitations}
\label{app:broader_limitation}
This paper addresses the benchmark prediction problem in scenarios with limited data. One potential limitation of our study is the relatively small number of models examined. For both the HELM-Lite and GLUE benchmarks, we have collected full benchmark results for fewer than 100 models. Despite conducting 100 random trials for each experiment, including additional and more diverse models could further strengthen the comprehensiveness and robustness of our analysis.

We do not anticipate any direct societal impacts from this work, such as potential malicious or unintended uses, nor do we foresee any significant concerns involving fairness, privacy, or security considerations. Additionally, we have not identified potential harms resulting from the application of this technology.

\end{document}